\documentclass[10pt,twocolumn]{article}
\usepackage[margin=1in]{geometry} 
\usepackage{times} 

\usepackage{indentfirst,csquotes}

\usepackage{amssymb,amsthm,amsmath}
\usepackage{booktabs}
\usepackage{xcolor,paralist,hyperref}

\usepackage[margin=1in]{geometry}
\usepackage{lipsum}
\usepackage{graphicx}
\usepackage{amsmath}
\usepackage{algorithm}
\usepackage{algpseudocode}
\usepackage{stfloats}
\usepackage{fontspec}

\usepackage[utf8]{inputenc} 
\usepackage[T1]{fontenc}    
\usepackage{amsfonts}       
\usepackage{nicefrac}       
\usepackage{microtype}      
\usepackage{xcolor}         
\definecolor{gaincolor}{RGB}{0, 155, 119}

\begin{document}
	
\title{\textbf{HyperVQ}: Enabling Hyperprior Entropy Modeling for VQ-Based Generative Image Compression}

\author{
	Niu Yi\\
	Xidian University\\
	\texttt{niuyi@mail.xidian.edu.cn}
	\and
	Xu Tianyi\\
	Xidian University\\
	\texttt{25171111417@stu.xidian.edu.cn}
	\and
	Ma Mingming\\
	Xidian University\\
	\texttt{mamingming@xidian.edu.cn}
	\and
	Wang Xinkun\\
	Xidian University\\
	\texttt{25171213993@stu.xidian.edu.cn}
}

\maketitle

\begin{abstract}
	Vector Quantization (VQ) based generative image compression has achieved remarkable perceptual quality. However, existing VQ codecs suffer from two fundamental limitations. First, they lack efficient content-adaptive entropy modeling and rely on static frequencies, leading to low coding efficiency. Second, the inherent conflict between discrete indices and continuous priors prevents true end-to-end joint Rate-Distortion (RD) optimization. 
	To resolve these issues, we propose HyperVQ, a principled framework that establishes a high-performance hyperprior entropy foundation for VQ-based codecs. The core insight of HyperVQ is to shift probability modeling entirely into the continuous embedding space. Instead of directly predicting probabilities for discrete symbols, HyperVQ predicts a high-dimensional continuous multivariate Gaussian distribution for the continuous latents. By treating the discrete codebook entries as fixed "anchors" in this space, we convert the continuous Gaussian density into categorical index probabilities based on relative distances. This elegant formulation provides a powerful, spatially-adaptive entropy engine and renders the cross-entropy rate objective fully differentiable, empowering the network to actively and dynamically optimize the RD trade-off during training. 
	To ensure practicality, we design the lightweight H Block and the Probability Estimation Engine (PEE) to facilitate highly parallel, millisecond-level inference. Experiments demonstrate that HyperVQ acts as a universal module across diverse VQ architectures (single-scale, large-codebook, RVQ), achieving an average bitrate saving of 18.5\%, which is 7.28$\times$ the saving achieved by conventional Huffman coding. This establishes a robust, RD-controllable foundation for next-generation generative image compression.
\end{abstract}

\section{Introduction}
\label{sec:intro}
In recent years, deep learning-based image compression has made significant progress~\cite{balle2016end}. End-to-end optimized autoencoders with powerful entropy models~\cite{balle2018variational, minnen2018joint} have become the dominant paradigm. Typically based on Scalar Quantization (SQ), they surpass classical codecs like VVC~\cite{bross2021overview} on pixel-level metrics (e.g., PSNR). However, optimizing for pixel-level metrics often leads to perceptually poor results, particularly at low bitrates.

To address this, Generative Image Compression (GIC) models (e.g., GANs~\cite{mentzer2020high}, VQ-VAEs~\cite{esser2021taming}) optimize the Rate-Distortion-Perception (RDP) trade-off, enabling highly realistic reconstructions at extremely low bitrates. Among these, Vector Quantization (VQ) based paradigms are particularly favored for discretizing features into semantically rich codebook indices.

Despite VQ's excellent performance in perceptual reconstruction, it faces a dual bottleneck that severely limits its potential. First, existing VQ-based codecs lack efficient, content-adaptive entropy modeling. By employing static frequency tables, they assume a stationary global distribution that ignores the recurring local patterns and spatial redundancies present in diverse image contents. Second, and more critically, the inherent conflict between discrete, unordered VQ indices and continuous priors prevents true end-to-end joint Rate-Distortion (RD) optimization. Because prior codecs lack a differentiable, content-adaptive probability model for the indices, the index-rate term remains unavailable during training. Consequently, the bitrate is often steered by manual heuristics (e.g., codebook size, downsampling factor) rather than actively optimized by the network.

\begin{figure*}[t]
	\vspace{-15pt}
	\centering
	\includegraphics[width=0.9\linewidth]{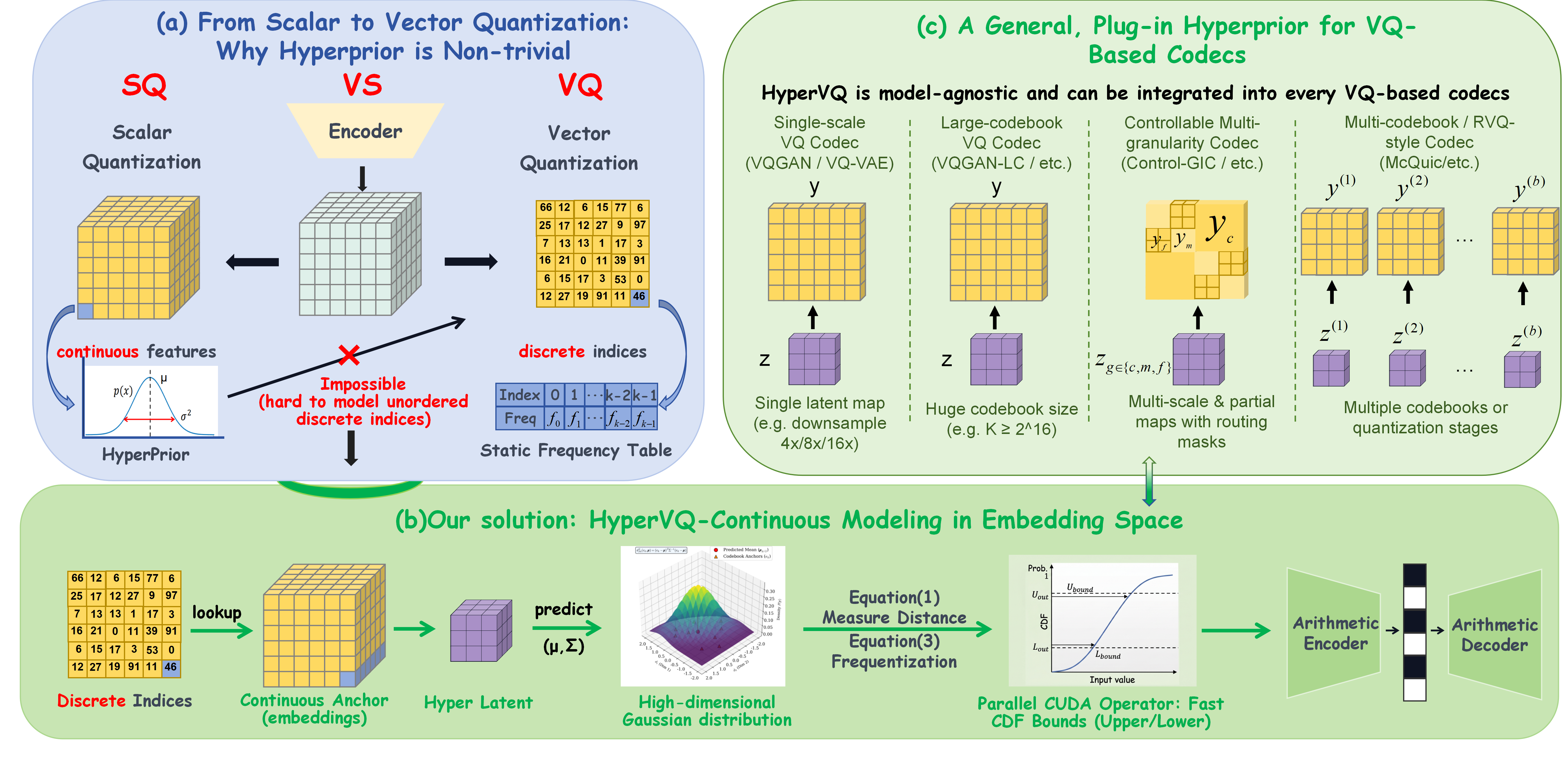}
	\caption{\textbf{HyperVQ: A Plug-and-Play Entropy Model and RD Optimizer for VQ-Based Codecs.} We present HyperVQ, which enables high-dimensional Gaussian modeling in the continuous embedding space of VQ indices to replace static frequency tables, bridging the gap between VQ and advanced entropy modeling. As a model-agnostic plug-in module, it supports every VQ-based codec variants out of the box.}
	\label{fig1:correlation}
	\vspace{-12pt}
\end{figure*}

In this paper, we introduce \textbf{HyperVQ}, a principled hyperprior entropy framework designed to resolve these bottlenecks. The core insight of HyperVQ lies in shifting the probability modeling entirely into the continuous embedding space. Instead of directly predicting probabilities for discrete symbols, HyperVQ predicts a high-dimensional continuous multivariate Gaussian distribution for the continuous latents at each spatial location. By treating the discrete codebook entries $\{\mathbf{e}_k\}$ as fixed ``anchors'' in this space, we mathematically convert this continuous Gaussian density into categorical index probabilities based on their relative distances (\S\ref{sec:method_principle}). This formulation not only provides a powerful, spatially-adaptive entropy engine, but crucially, it renders the cross-entropy rate objective fully differentiable. By establishing this differentiable gradient force field in the embedding space, HyperVQ empowers the network to actively and dynamically optimize the RD trade-off during training, effectively shifting RD scheduling from hand-tuned heuristics to the network itself.

As a universal, plug-and-play module, HyperVQ seamlessly integrates across diverse VQ scenarios, including single-scale, large-codebook, residual quantization, and multi-granularity settings (\S\ref{sec:exp_generalization}). By successfully adapting to varied structural constraints, including the entropy modeling of hierarchical scales and the handling of partial feature maps caused by arbitrary rate control (\S\ref{sec:method_instantiation}), HyperVQ rigorously proves its adaptive capability and potential to serve as a foundational component for the entire VQ family. 

Our main contributions are summarized as follows:
\begin{itemize}
	\item We propose \textbf{HyperVQ}, the first principled hyperprior entropy model that resolves the fundamental conflict between continuous priors and discrete symbols by shifting probability modeling entirely into the continuous embedding space.
	\item We mathematically derive a fully differentiable cross-entropy rate objective that maps continuous Gaussian densities to discrete categorical probabilities. This establishes an elegant gradient force field that brings true end-to-end, dynamic Rate-Distortion (RD) optimization to VQ-based generative compression.
	\item To ensure practicality, we design the lightweight \textbf{H Block} and the \textbf{Probability Estimation Engine (PEE)} to facilitate highly parallel, millisecond-level inference (\S\ref{sec:method_design}). We demonstrate that HyperVQ acts as a universal module, achieving an average bitrate saving of 18.5\%, which is 7.28$\times$ the saving achieved by conventional Huffman coding.
	\item We instantiate HyperVQ on the most demanding controllable multi-granularity setting, termed \textbf{HyperVQ-Controllable Generative Image Compression (HVQ-CGIC)}, achieving state-of-the-art performance across multiple metrics; notably, it obtains over 61\% bit reduction on Kodak while maintaining the same LPIPS as the previous SOTA (\S\ref{sec:exp_hardest_case}).
\end{itemize}

\section{Related Work}
\label{sec:Related_Work}

\subsection{Learned image compression}
Grounded in Shannon's rate-distortion theory~\cite{shannon1959coding}, Ballé et al.~\cite{balle2016end} established the dominant paradigm of end-to-end optimization based on Scalar Quantization (SQ). Subsequent research has advanced this paradigm by exploring powerful architectures and accurate probability models.

Architectural advancements include sophisticated nonlinear transforms~\cite{rippel2017real, agustsson2019generative, johnston2018improved, li2018learning}, content-adaptive mechanisms~\cite{shin2022expanded, pan2022content, li2022content}, and Transformer~\cite{zhu2022transformer, liu2023learned} or Diffusion-based models~\cite{yang2023lossy, careil2023towards}.

For entropy modeling, the Hyperprior~\cite{balle2018variational} parameterizes a conditional distribution $p(\hat{\mathbf{y}}|\mathbf{z})$ for continuous latents. Combined with autoregressive priors~\cite{mentzer2018conditional, minnen2018joint, lee2018context}, these learned methods finally outperformed BPG in PSNR, inspiring numerous subsequent R-D improvements~\cite{choi2019variable, cheng2020learned, lin2020spatial, minnen2020channel, mentzer2020high, zhang2021attention, he2021checkerboard, yang2021slimmable, gao2021neural, kim2022joint, zhang2022multi, he2022elic, jiang2023mlic, wang2023evc, jia2024generative}.

While effective, these foundational works are almost exclusively designed for SQ. Vector Quantization (VQ) offers superior theoretical rate-distortion properties but has been less common. Beyond early soft-to-hard VQ~\cite{agustsson2017soft} and recent Lattice VQ (LVQ) frameworks~\cite{zhang2023lvqac, feng2023nvtc, lei2024approaching, zhang2024learning}, effectively modeling the entropy of discrete, unordered VQ indices remains a significant gap.

\subsection{VQ-Based generative image compression}
Recently, VQ-based generative compression has evolved to optimize the Rate-Distortion-Perception (RDP) trade-off across diverse architectures. These include standard single-scale VQ codecs~\cite{esser2021taming, zheng2022movq}, large-codebook designs for higher fidelity~\cite{zhu2024scaling}, controllable multi-granularity codecs for variable bitrates~\cite{li2024once}, and multi-codebook or Residual VQ (RVQ) models utilizing cascading quantization stages~\cite{zhu2022unified}.

Across all these families, efficient entropy modeling remains a critical bottleneck. Existing codecs estimate index entropy using static, global frequency tables, failing to exploit content-specific recurring patterns and strong spatial redundancies. We address this fundamental gap by shifting probability modeling entirely into the continuous embedding space. HyperVQ introduces a general, codebook-size-independent hyperprior entropy model serving as a universal plug-in for all aforementioned VQ families. It enables true end-to-end Rate-Distortion optimization via a fully differentiable cross-entropy objective, completely bypassing the parameter explosion of discrete classification.

\begin{figure*}[t]
	\vspace{-8pt}
	\centering
	\includegraphics[width=1.0\linewidth]{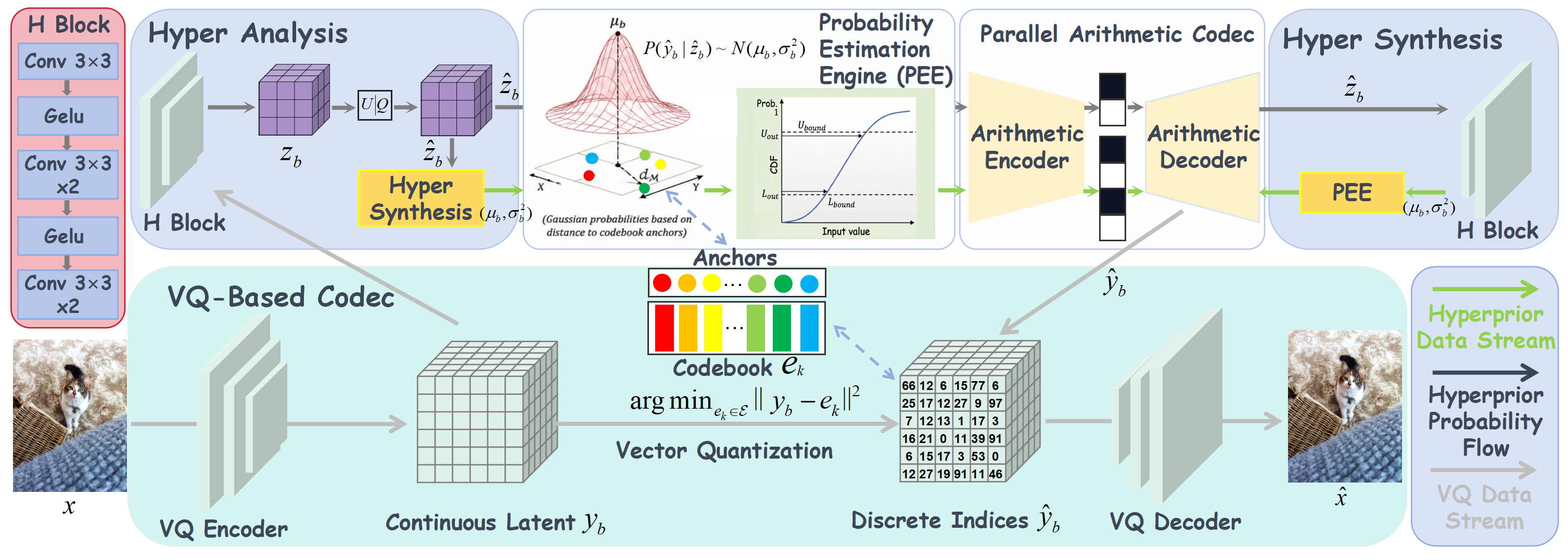}
	\caption{Overall framework of HyperVQ. The architecture consists of a base VQ codec, a generic HyperVQ module that predicts embedding-space distributions, and a parallel arithmetic codec. The Probability Estimation Engine (PEE) converts the predicted continuous density into discrete frequencies for efficient transmission.}
	\label{fig:framework}
	\vspace{-15pt}
\end{figure*}

\section{Method}
\label{sec:Method}

\subsection{Framework overview}
\label{sec:method_overview}
In this section, we present HyperVQ, a general entropy modeling framework. As illustrated in Figure~\ref{fig:framework}, HyperVQ is designed as a universal component that can be integrated into diverse vector quantization (VQ) based codecs. Our design aims to resolve the inefficiency of static frequency tables by introducing content-adaptive, learned entropy modeling. The framework consists of three core components.

1) Base VQ backbone. This part establishes the system foundation, using a pre-trained or jointly trained encoder to map the input image $x$ into discrete codebook indices across specified feature branches $b \in \mathcal{B}$.
2) HyperVQ module. Serving as the analytical core, this module predicts content-adaptive probability distributions for the VQ indices by operating within the continuous embedding space, bridging the gap between continuous hyperpriors and discrete symbols.
3) Parallel arithmetic codec. To ensure inference efficiency, we develop an accelerated engine that utilizes custom CUDA kernels to perform parallel density-to-probability mapping and arithmetic coding.

\subsection{Principle of hyperprior for VQ indices}
\label{sec:method_principle}

\subsubsection{Problem Setting: Bridging continuous priors and discrete Indices}
The success of hyperpriors in Scalar Quantization (SQ)~\cite{balle2018variational, minnen2018joint} arises from predicting parameters $(\mu, \sigma)$ for a continuous latent variable. However, VQ indices are discrete and unordered symbols, lacking the continuous spatial structure inherent in SQ latents. Consider an input image $x \in \mathbb{R}^{3 \times H \times W}$. A generic VQ encoder outputs continuous feature maps $\{y_b\}_{b \in \mathcal{B}}$, where $b$ denotes a specific branch, scale, or quantization stage. This multi-branch formulation enables HyperVQ to support diverse VQ architectures, ranging from multi-granularity (multi-scale) representations to multi-codebook (e.g., residual) quantization stages. The VQ step maps the $D$-dimensional feature vector $(y_b)_{ij}$ at each spatial location $(i,j)$ to a discrete index $(\hat{y}_b)_{ij} \in \{1, \ldots, K\}$. Constructing a content-adaptive probability model $P(\hat{y}_b \mid \hat{z}_b)$ for these indices is the critical challenge addressed by HyperVQ.

\subsubsection{Core Idea: Gaussian prior in the embedding Space}
Our key insight is that although the indices $(\hat{y}_b)_{ij}$ are discrete, their corresponding codebook embedding vectors $\mathbf{e}_k \in \mathbb{R}^D$ reside in a continuous $D$-dimensional embedding space. Consequently, rather than modeling the index $k$ directly, we predict a conditional probability distribution within this continuous embedding space. 

Specifically, for each branch $b \in \mathcal{B}$, a hyperprior branch processes the feature $y_b$ to generate a continuous hyper-latent $z_b$. After quantization to $\hat{z}_b$, a HyperSynthesis network predicts the parameters of a multivariate Gaussian distribution, denoted as $\Theta_{b,ij}(\hat{z}_b) = (\boldsymbol{\mu}_{b,ij}, \boldsymbol{\Sigma}_{b,ij})$. Here, $\boldsymbol{\mu}_{b,ij} \in \mathbb{R}^D$ represents the expected feature vector (expected content) at that location, and $\boldsymbol{\Sigma}_{b,ij} \in \mathbb{R}^{D \times D}$ captures the prediction uncertainty.

\begin{figure*}[t]
	\vspace{-10pt}
	\centering
	\includegraphics[width=1.0\linewidth]{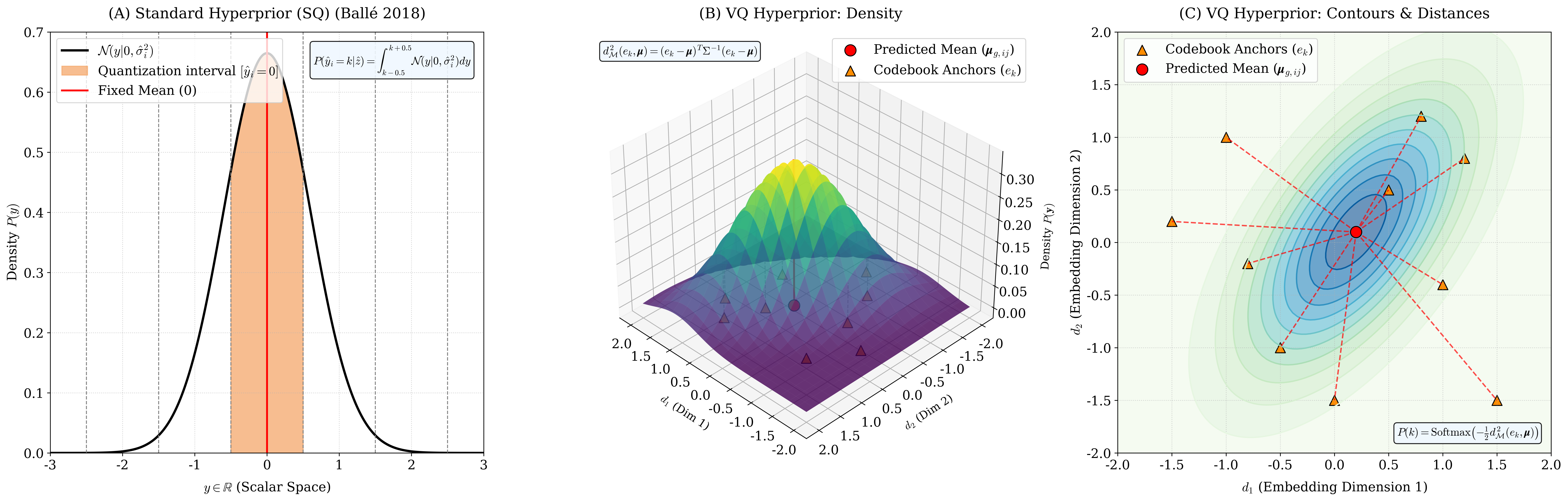}
	\caption{(a) Standard scalar quantization (SQ) hyperprior. (b)(c) Our proposed HyperVQ models a Gaussian probability cloud in the $D$-dim embedding space. The codebook entries act as anchors, and their probabilities are determined by their relative distances to the predicted distribution center.}
	\label{fig:principles}
	\vspace{-10pt}
\end{figure*}

\subsubsection{From continuous Gaussian to discrete probabilities via Mahalanobis Distance}
\label{sec:Math}
The next step is to bridge the continuous Gaussian parameters $(\boldsymbol{\mu}_{b,ij}, \boldsymbol{\Sigma}_{b,ij})$ to the probabilities $P((\hat{y}_b)_{ij}=k \mid \hat{z}_b)$ of the discrete indices $k$. We treat the codebook embeddings $\mathbf{e}_k$ as fixed ``anchors'' in the embedding space. As illustrated in Figure~\ref{fig:principles}, we hypothesize that the probability of selecting index $k$ at location $(i,j)$ is related to the squared Mahalanobis Distance between its anchor $\mathbf{e}_k$ and the predicted Gaussian distribution $\mathcal{N}(\boldsymbol{\mu}_{b,ij}, \boldsymbol{\Sigma}_{b,ij})$, expressed as
\begin{equation}
	d^2_{\mathcal{M}}(\mathbf{e}_k, \boldsymbol{\mu}_{b,ij}) = (\mathbf{e}_k - \boldsymbol{\mu}_{b,ij})^{\top} (\boldsymbol{\Sigma}_{b,ij})^{-1} (\mathbf{e}_k - \boldsymbol{\mu}_{b,ij})
\end{equation}
The categorical probability distribution is obtained via a Softmax function as follows
\begin{equation}
	P((\hat{y}_b)_{ij}=k \mid \hat{z}_b) = \frac{\exp\left( -\frac{1}{2} d^2_{\mathcal{M}}(\mathbf{e}_k, \boldsymbol{\mu}_{b,ij}) \right)}{\sum_{\ell=1}^{K} \exp\left( -\frac{1}{2} d^2_{\mathcal{M}}(\mathbf{e}_\ell, \boldsymbol{\mu}_{b,ij}) \right)}
\end{equation}
For computational efficiency and stability, we assume the covariance matrix is diagonal and isotropic, i.e., $\boldsymbol{\Sigma}_{b,ij} = (\sigma_{b,ij})^2 \mathbf{I}$. Under this simplification, the probability reduces to the scaled squared Euclidean distance, given by
\begin{equation}
	P((\hat{y}_b)_{ij}=k \mid \hat{z}_b) = \frac{\exp\left( -\frac{\|\mathbf{e}_k - \boldsymbol{\mu}_{b,ij}\|^2}{2(\sigma_{b,ij})^2} \right)}{\sum_{\ell=1}^{K} \exp\left( -\frac{\|\mathbf{e}_\ell - \boldsymbol{\mu}_{b,ij}\|^2}{2(\sigma_{b,ij})^2} \right)}
	\label{eq:hypervq_prob}
\end{equation}
Intuitively, while SQ hyperpriors predict a 1D Gaussian, our method predicts a high-dimensional Gaussian probability ``cloud'' in the embedding space. Codebook anchors closer to the cloud's center receive higher probability, while the spread $\sigma_{b,ij}$ reflects the model's uncertainty, effectively controlling the concentration of the distribution and the resulting bitrate.

\subsubsection{Implications for efficient entropy coding and RD optimization}
The core value of Equation~\eqref{eq:hypervq_prob} lies in providing a complete categorical probability distribution $P_{b,ij} = \{p_{b,ij,k}\}_{k=1}^K$ over the VQ index $(\hat{y}_b)_{ij}$ at $(i,j)$. According to information theory, arithmetic coding a symbol using this distribution achieves an expected code length close to its Shannon entropy. The expected bitrate for encoding all VQ indices at branch $b$ is formulated as
\begin{align}
	R_{\hat{y}_b} &= \mathbb{E}_{x} \left[ \sum_{i,j} H(P_{b,ij}^{\star}, P_{b,ij}) \right] \nonumber \\
	&= \mathbb{E}_{x} \left[ \sum_{i,j} \left( H(P_{b,ij}^{\star}) + D_{\text{KL}}(P_{b,ij}^{\star} \parallel P_{b,ij}) \right) \right]
	\label{eq:rate_cross_entropy}
\end{align}
where $P_{b,ij}^{\star}$ denotes the real but unknown true distribution of the index given image $x$. Crucially, by modeling this true discrete distribution (derived from the deterministic nearest-neighbor VQ assignment) as a Dirac delta function, we directly minimize this cross-entropy via backpropagation. This mathematical formulation elegantly establishes a differentiable gradient force field in the embedding space (fully derived in Appendix~\ref{sec:supp_theory}). Specifically, the gradients actively pull the predicted mean $\boldsymbol{\mu}_{b,ij}$ toward the target anchor while repelling it from other anchors based on their assigned probabilities. This empowers the network to dynamically optimize the RD trade-off during training.

\subsection{Module design and parallel implementation}
\label{sec:method_design}
This section maps the VQ hyperprior principles onto the network modules of HyperVQ, following the overall pipeline illustrated in Figure~\ref{fig:framework}.

\subsubsection{Base VQ backbone and HyperVQ branches}
A generic VQ encoder distills hierarchical representations from the input image $x$ into continuous latents $\{y_b\}_{b \in \mathcal{B}}$. The VQ module maps each continuous feature vector to its nearest entry in a shared codebook $\mathcal{E}$, producing discrete index maps $\hat{y}_b$. For each branch $b \in \mathcal{B}$, an independent hyperprior branch consisting of a HyperAnalysis network $H_a$ compresses $y_b$ into a hyper-latent $z_b$. During training, $z_b$ is quantized via additive uniform noise $\hat{z}_b = z_b + \mathcal{U}(-0.5, 0.5)$; during inference, we use rounding $\hat{z}_b = \text{round}(z_b)$. Subsequently, a HyperSynthesis network $H_s$ generates the Gaussian parameters $(\boldsymbol{\mu}_b, \sigma_b) = H_s(\hat{z}_b)$.

\subsubsection{Parallel probability estimation and entropy coding}
A defining feature of our framework is its inference efficiency, which is rooted in the spatial independence of the index probabilities. Given the isotropic assumption, the categorical distribution for each spatial location $(i,j)$ depends only on the locally predicted parameters $(\boldsymbol{\mu}_{b,ij}, \sigma_{b,ij})$. This property enables massive parallelization across the feature map.

First, the Probability Estimation Engine (PEE) computes the frequencies simultaneously across all locations by parallelizing the distance measurements to the codebook anchors. Second, to interact with the arithmetic coder, the PEE does not need to transmit the entire probability table sequentially. Instead, for a target index $k$ at location $(i,j)$, it computes and transmits the lower and upper bounds of the Cumulative Distribution Function (CDF) as follows
\begin{equation}
	[F_{\text{lower}}, F_{\text{upper}}] = \left[ \sum_{\ell=1}^{k-1} P((\hat{y}_b)_{ij}=\ell \mid \hat{z}_b), \sum_{\ell=1}^{k} P((\hat{y}_b)_{ij}=\ell \mid \hat{z}_b) \right]
\end{equation}
By framing the problem this way, we implement a custom CUDA kernel to parallelize this CDF bound construction and the subsequent interval updates in the arithmetic codec, ensuring high-speed encoding and decoding without sequential bottlenecks (see Appendix~\ref{sec:supp_algo} for the algorithmic pseudo-code and latency micro-benchmarks).

\subsubsection{Instantiation on a Controllable Multi-Granularity Codec}
\label{sec:method_instantiation}

We evaluate the robustness of HyperVQ by instantiating it on a controllable multi-granularity codec, Control-GIC~\cite{li2024once}, which presents a highly demanding latent environment (the complete architectural flowchart is illustrated in Appendix~\ref{sec:supp_arch}). The primary challenge stems from its arbitrary rate control mechanism that dynamically routes features across hierarchical scales via spatial routing masks. Note that although these spatial routing masks are not entropy-coded, their bit overhead is properly calculated and included in the total reported bitrate.

Unlike traditional VQ codecs that process complete and spatially uniform feature maps, this routing mechanism yields sparse and fragmented latent representations. Each granularity level contains only partial information and exhibits non-contiguous spatial distributions that shift based on content complexity. Such irregularity poses a significant obstacle for entropy modeling because standard methods rely on dense and stable spatial structures to predict probabilities reliably.

HyperVQ addresses these challenges through a unified hyperprior modeling strategy across all granularities. Since the framework utilizes a shared codebook, our module can effectively parameterize the categorical distributions of these sparse indices by leveraging the aggregated multi-scale context. This enables the framework to maintain high efficiency despite the fragmented nature of the latent space. The Rate-Distortion performance for this challenging instantiation is detailed in the experimental section to demonstrate the robustness and flexibility of our method.

\subsection{Objective and optimization}
\label{sec:loss_function}
The entire framework is jointly optimized to minimize a unified Rate-Distortion (RD) objective. By introducing the differentiable cross-entropy rate term derived in \S\ref{sec:method_principle}, HyperVQ allows the network to explicitly balance reconstruction quality and bitrate consumption according to the following formulation
\begin{align}
	\mathcal{L} &= \mathbb{E}[D(x,\hat{x})] + \sum_{b \in \mathcal{B}} \left( \lambda_{y,b} R_{\hat{y}_b} + \lambda_{z,b} R_{\hat{z}_b} \right) \label{eq:unified_loss} \\
	D(x,\hat{x}) &= \mathcal{L}_{\text{rec}} + \mathcal{L}_{\text{percep}} + \lambda_{\text{GAN}}\mathcal{L}_{\text{GAN}} + \lambda_{\text{VQ}}\mathcal{L}_{\text{VQ}}
\end{align}
where $\mathcal{L}_{\text{rec}} = \|x - \hat{x}\|_2^2$ is the MSE reconstruction loss, and $\mathcal{L}_{\text{percep}}$ is the LPIPS perceptual loss. The adversarial loss is defined as
\begin{equation}
	\mathcal{L}_{\text{GAN}} = -\log \mathcal{D}(\hat{x})
\end{equation}
where $\mathcal{D}$ is a PatchGAN discriminator. To stabilize the embedding space, the VQ commitment loss is formulated across all branches $b \in \mathcal{B}$ by
\begin{equation}
	\mathcal{L}_{\text{VQ}} = \sum_{b \in \mathcal{B}} \left( \|\text{sg}[y_b] - \mathbf{e}_{\hat{y}_b}\|_2^2 + \beta \|\text{sg}[\mathbf{e}_{\hat{y}_b}] - y_b\|_2^2 \right)
\end{equation}
where $\text{sg}[\cdot]$ denotes the stop-gradient operator, $\mathbf{e}_{\hat{y}_b}$ represents the quantized features via codebook lookup. 

The rate term consists of the cross-entropy for VQ indices ($R_{\hat{y}_b}$) and the bitrate of hyper-latents ($R_{\hat{z}_b}$). The bitrate weights $\lambda_{y,b}$ and $\lambda_{z,b}$ control the exact RD curve positioning. Notably, this formulation hands over the RD scheduling directly to the gradient descent process, enabling the VQ backbone and the hyperprior to co-adapt perfectly for optimal rate-distortion-perception (RDP) trade-offs. Implementation details and the progressive joint-training protocol are provided in Appendix~\ref{sec:supp_training}.

\section{Experiments}
\label{sec:Experiments}

\subsection{Experimental protocol}
\label{sec:exp_protocol}

\textbf{Evaluated Codecs.} To evaluate generalizability, we test HyperVQ across diverse representative architectures. These include single-scale codecs (VQGAN~\cite{esser2021taming}, Fine-tuned VQ~\cite{mao2024extreme}, Mo-VQGAN~\cite{zheng2022movq}), large-codebook models (VQGAN-LC~\cite{zhu2024scaling}), controllable multi-granularity codecs (Control-GIC~\cite{li2024once}), and multi-codebook RVQ frameworks (McQuic~\cite{zhu2022unified}). 

\textbf{Datasets and Metrics.} Model evaluation utilizes ImageNet-1K Val~\cite{russakovsky2015imagenet} and the standard Kodak~\cite{kodak_dataset} dataset for comprehensive benchmarking. Additional high-resolution results on DIV2K~\cite{agustsson2017ntire} and the CLIC 2020 professional set~\cite{toderici2020clic} are provided in Appendix~\ref{sec:supp_quantitative}. To comprehensively assess both objective fidelity and subjective realism, we measure bitstream size via bits per pixel (bpp) and employ a robust suite of metrics including LPIPS and DISTS for reference-based perceptual quality, along with FID, NIQE, and IS for no-reference generative realism, and PSNR for pixel-level distortion.

\textbf{Implementation Details.} Training configurations including optimizers and schedules follow the original papers for each respective backbone. Detailed integration protocols for equipping them with HyperVQ are provided in Appendix~\ref{sec:supp_training}. 

\begin{table*}[t]
	\centering
	\caption{Bitrate savings of HyperVQ across different VQ-based models. Gain is computed relative to the theoretical bpp. For Residual Vector Quantization (RVQ) baselines, stages are cumulatively combined (+).}
	\label{table:cross_model}
	\resizebox{0.95\linewidth}{!}{
		\begin{tabular}{ccccccc}
			\toprule
			\textbf{Dataset} & \textbf{Method} & \textbf{Codebook} & \textbf{Scale} & \textbf{Theoretical bpp} & \textbf{Huffman (Gain)} & \textbf{HyperVQ (Gain)} \\
			\midrule
			ImageNet-1K Val & VQGAN-LC (NeurIPS'24) & 16384 & $16\times$ & 0.0547 & 0.0541 (\textcolor{blue}{1.1\%}) & 0.0458 (\textcolor{teal}{16.3\%}) \\
			&  & 100K & $16\times$ & 0.0641 & 0.0642 (\textcolor{red}{-0.2\%}) & 0.0564 (\textcolor{teal}{12.0\%}) \\
			&  & 100K & $8\times$ & 0.2595 & 0.2516 (\textcolor{blue}{3.0\%}) & 0.2072 (\textcolor{teal}{20.2\%}) \\
			\cmidrule{2-7}
			& VQGAN (CVPR'21) & 1024 & $4\times$ & 0.6250 & 0.5844 (\textcolor{blue}{6.5\%}) & 0.4703 (\textcolor{teal}{24.8\%}) \\
			&  & 2048 & $4\times$ & 0.6875 & 0.6473 (\textcolor{blue}{5.8\%}) & 0.5119 (\textcolor{teal}{25.5\%}) \\
			\midrule
			Kodak & Fine-tuned VQ (DCC'24) & 1024 & $4\times$ & 0.6250 & 0.6040 (\textcolor{blue}{3.4\%}) & 0.5621 (\textcolor{teal}{10.1\%}) \\
			&  & 1024 & $8\times$ & 0.1563 & 0.1522 (\textcolor{blue}{2.6\%}) & 0.1335 (\textcolor{teal}{14.6\%}) \\
			&  & 1024 & $16\times$ & 0.0391 & 0.0389 (\textcolor{blue}{0.5\%}) & 0.0337 (\textcolor{teal}{13.8\%}) \\
			\cmidrule{2-7}
			& Mo-VQGAN (NeurIPS'22) & 1024 & $4\times$ & 0.6250 & 0.5947 (\textcolor{blue}{4.8\%}) & 0.4984 (\textcolor{teal}{20.3\%}) \\
			\cmidrule{2-7}
			& Control-GIC (ICLR'25) & 1024 & $4\times$ & 0.6250 & 0.5943 (\textcolor{blue}{4.9\%}) & 0.4930 (\textcolor{teal}{21.1\%}) \\
			&  & 1024 & $8\times$ & 0.1562 & 0.1533 (\textcolor{blue}{1.9\%}) & 0.1381 (\textcolor{teal}{11.6\%}) \\
			&  & 1024 & $16\times$ & 0.0391 & 0.0389 (\textcolor{blue}{0.5\%}) & 0.0327 (\textcolor{teal}{16.4\%}) \\
			\cmidrule{2-7}
			& McQuic (CVPR'22) & $2 \times 8192$ & $16\times$ & 0.1016 & 0.1005 (\textcolor{blue}{1.1\%}) & 0.0766 (\textcolor{teal}{24.6\%}) \\
			& & $+ 2 \times 2048$ & $ + 32\times$ & 0.1230 & 0.1216 (\textcolor{blue}{1.1\%}) & 0.0945 (\textcolor{teal}{23.2\%}) \\
			& & $+ 2 \times 512$ & $ + 64\times$ & 0.1277 & 0.1263 (\textcolor{blue}{1.1\%}) & 0.0983 (\textcolor{teal}{23.0\%}) \\
			\midrule
			\multicolumn{5}{r}{\textbf{Average Gain}} & \textbf{\textcolor{blue}{2.54\%}} & \textbf{\textcolor{teal}{18.5\%}} \\
			\bottomrule
		\end{tabular}
	}
\end{table*}

\subsection{Cross-Model generalization across VQ-Based codecs}
\label{sec:exp_generalization}

To verify universal applicability, we integrate HyperVQ into established VQ codecs, comparing the resulting coding efficiency against theoretical bitrate and Huffman coding.

As demonstrated in Table~\ref{table:cross_model}, HyperVQ consistently yields significant bitrate savings across diverse architectures, codebook sizes, and scales. HyperVQ effectively mines latent redundancies by leveraging high-dimensional context within the embedding space (corroborated in Appendix~\ref{sec:supp_entropy_bit}). This content-adaptive modeling particularly benefits large-codebook and complex hierarchical paradigms, validating HyperVQ as a plug-and-play module for the entire VQ family.

\subsection{Detailed results of HVQ-CGIC}
\label{sec:exp_hardest_case}

We further analyze our framework's most challenging instantiation, namely the controllable multi-granularity setting (HVQ-CGIC). Unlike standard codecs, this setting dynamically routes features to generate multi-scale and partial feature maps. Modeling the entropy of these sparse indices rigorously tests the hyperprior's robustness.

\begin{figure}[htbp] 
	\centering
	\includegraphics[width=0.5\textwidth]{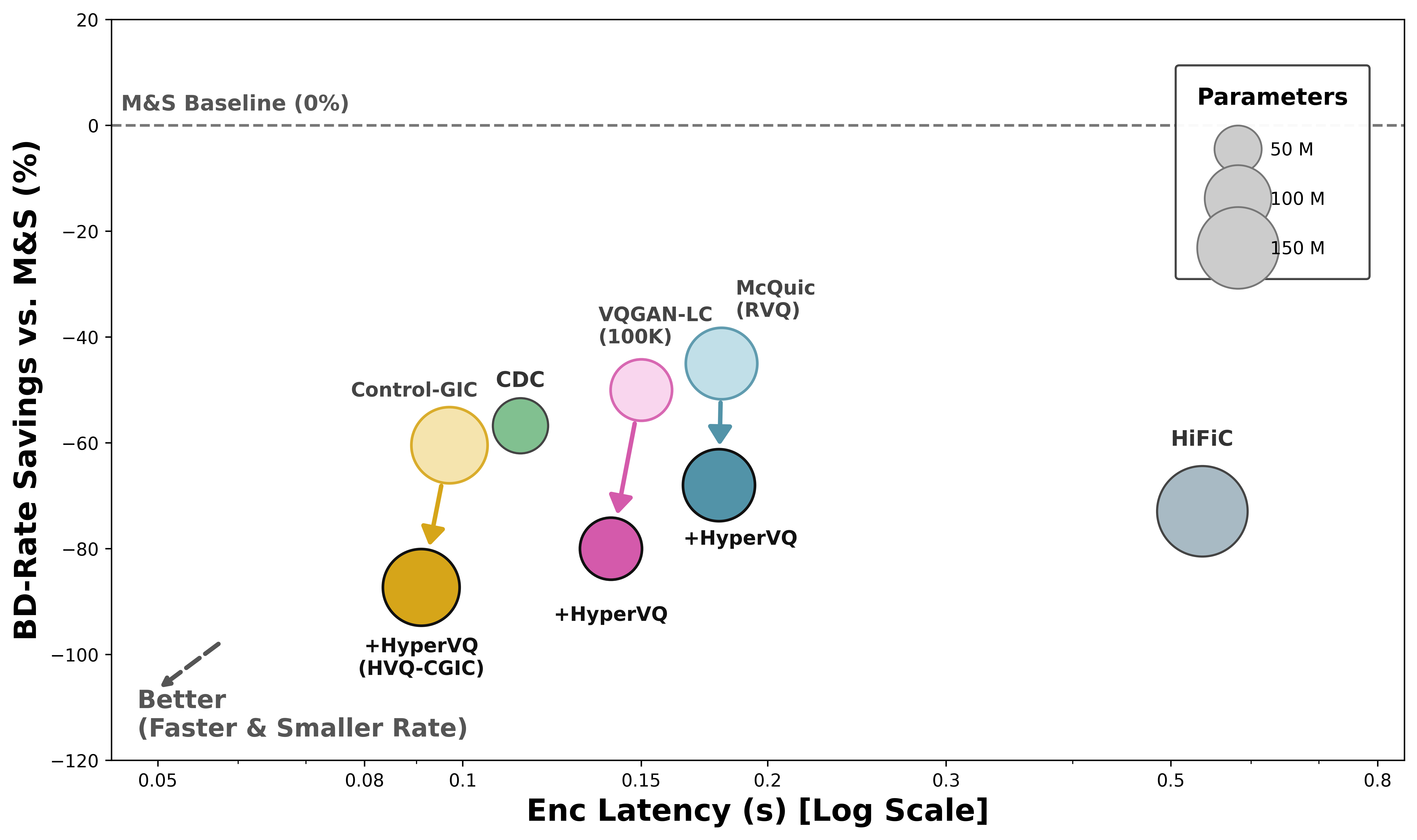}
	\caption{Visualizing the efficiency-performance trade-off. Bubble size represents parameter count. Our HyperVQ significantly pushes VQ-based codecs toward the ``Better'' quadrant (lower-left), achieving superior RD performance with negligible latency overhead.}
	\label{fig:efficiency_bubble}
\end{figure}

\textbf{Baseline Methods.} We compare our method against traditional codecs (BPG~\cite{bellard2018bpg}, VVC~\cite{bross2021overview}), classical neural models (M\&S~\cite{balle2018variational}), variable-rate/progressive frameworks (SCR~\cite{lee2022selective}, CTC~\cite{jeon2023context}), and state-of-the-art generative approaches including HiFiC~\cite{mentzer2020high} and CDC~\cite{yang2023lossy}. Additionally, expanded comparisons with recent non-VQ continuous-latent generative paradigms, such as MS-ILLM~\cite{muckley2023improving}, CRDR~\cite{iwai2024controlling}, and EGIC~\cite{korber2024egic}, are provided in Appendix~\ref{sec:supp_non_vq_comp}.

\begin{figure}[t]
	\centering
	\includegraphics[width=1.0\linewidth]{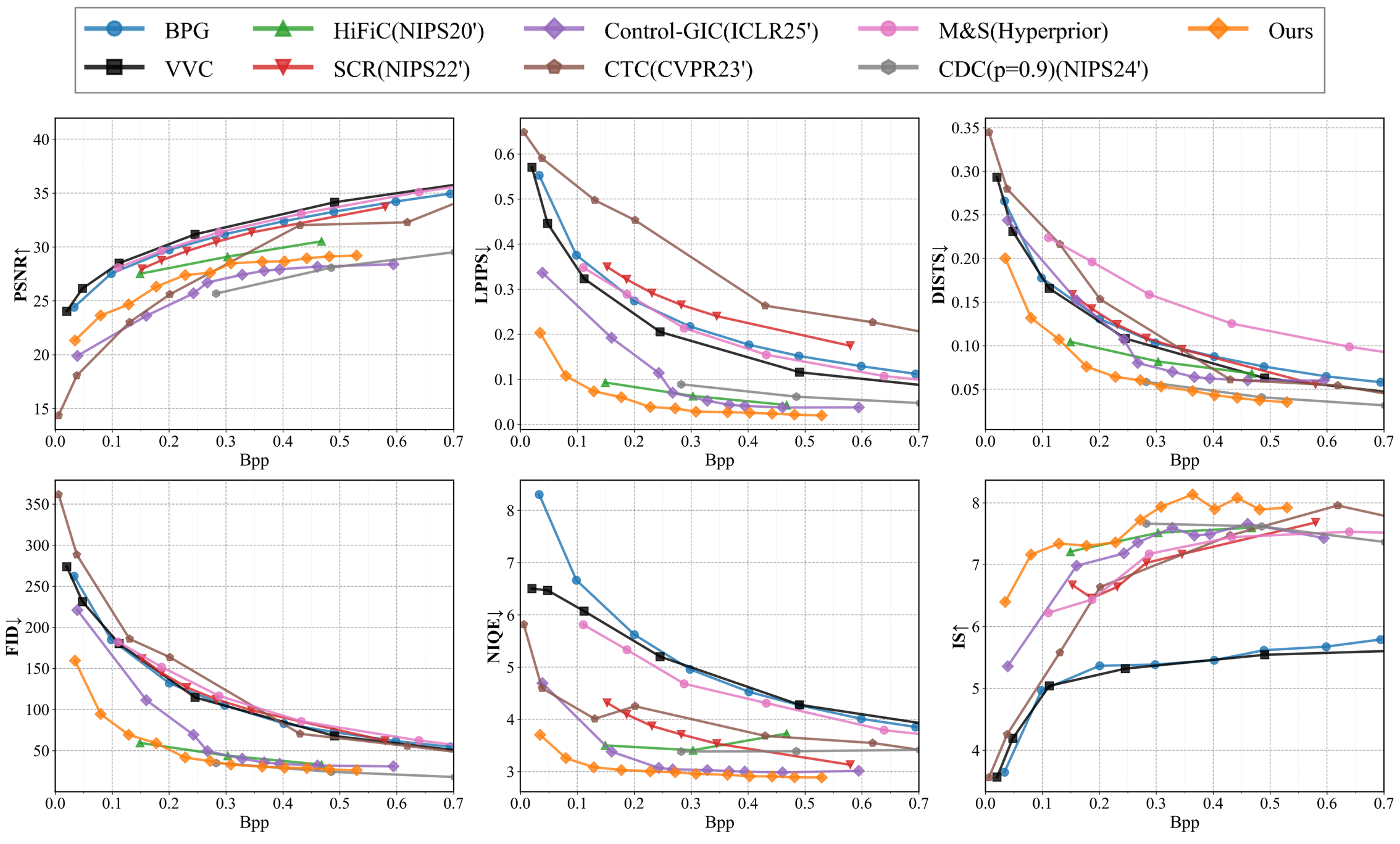}
	\vspace{-15pt}
	\caption{Rate-distortion and perceptual comparisons on Kodak. HVQ-CGIC pushes the Pareto frontier across multiple metrics.}
	\label{fig:rd_curve}
	\vspace{-15pt}
\end{figure}

\begin{figure}[htbp]
	\centering
	\includegraphics[width=1.0\linewidth]{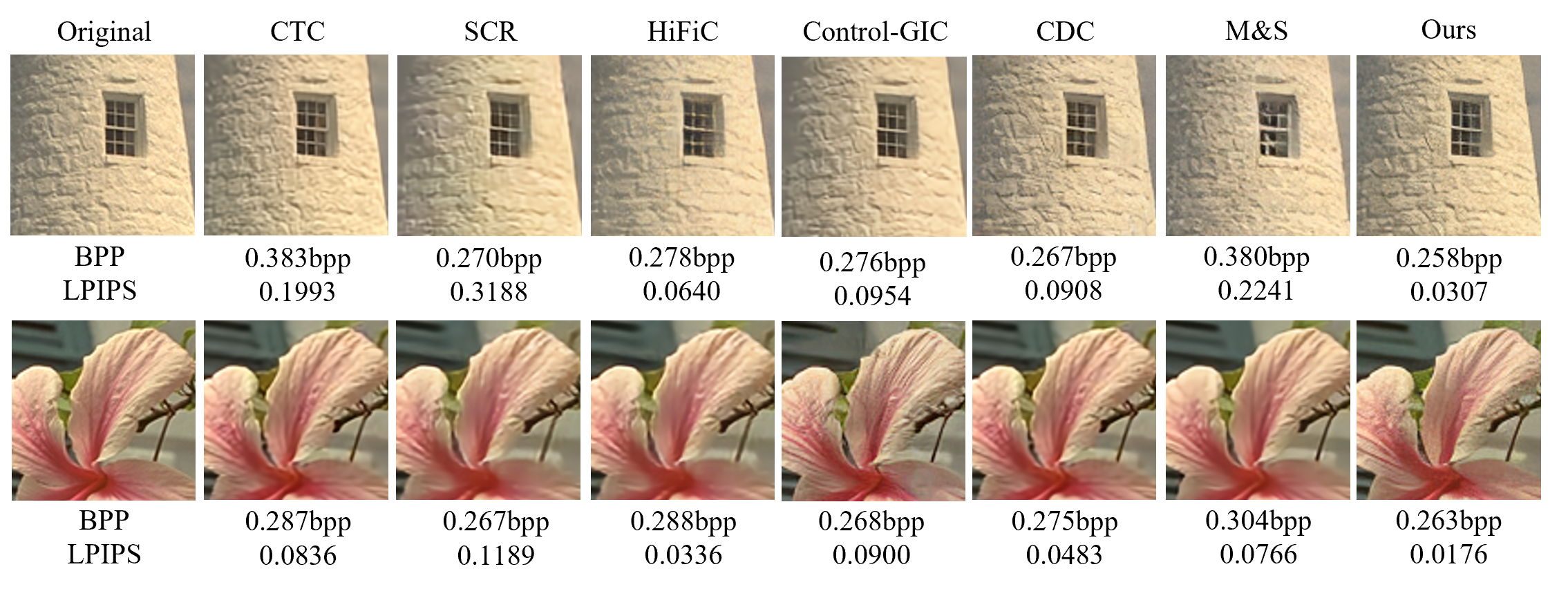}
	\vspace{-15pt}
	\caption{Qualitative comparisons of reconstruction fidelity between HVQ-CGIC and baseline methods at extremely low bitrates.}
	\label{fig:qualitative_show}
	\vspace{-10pt}
\end{figure}

\textbf{Quantitative and Qualitative Performance.} Figure~\ref{fig:rd_curve} shows HVQ-CGIC significantly outperforms baselines across perceptual metrics (LPIPS, NIQE, DISTS, FID, IS). Visual comparisons in Figure~\ref{fig:qualitative_show} confirm it preserves sharper structures and more realistic textures at highly constrained bitrates (see Appendix~\ref{sec:supp_visuals} for extended comparisons).

\textbf{Efficiency and BD-Rate Evaluation.} Table~\ref{tab:bd_rate_efficiency} details the trade-off between performance gains and computational cost. HVQ-CGIC achieves the highest BD-Rate savings against the M\&S anchor while maintaining highly competitive inference speeds. 

\begin{table*}[t]
	\centering
	\caption{Comprehensive Comparison of Efficiency and RD Performance. 
		\textit{Params} indicates the number of network parameters in Millions. 
		\textit{Time} reflects the average Encoding/Decoding latency in Seconds (s). 
		\textit{BD-Rate Savings (\%)} are calculated relative to M\&S \cite{balle2018variational}. 
		\textbf{Bold} indicates the best result, \underline{underline} indicates the second best, and N/A indicates data not available or applicable.}
	\label{tab:bd_rate_efficiency}
	\resizebox{0.95\textwidth}{!}{%
		\begin{tabular}{l|ccc|ccccc}
			\toprule
			\textbf{Method} & \multicolumn{3}{c|}{\textbf{Efficiency}} & \multicolumn{5}{c}{\textbf{BD-Rate Savings (\%) of Ours vs. Method}} \\
			\cmidrule(lr){2-4} \cmidrule(lr){5-9}
			& \textbf{Params (M)} & \textbf{Enc. (s)} & \textbf{Dec. (s)} & \textbf{LPIPS} $\downarrow$ & \textbf{DISTS} $\downarrow$ & \textbf{FID} $\downarrow$ & \textbf{NIQE} $\downarrow$ & \textbf{IS} $\downarrow$ \\
			\midrule
			\textbf{Traditional Codecs} & & & & & & & & \\
			BPG \cite{bellard2018bpg} & N/A & 0.196 & 0.342 & 4.51 & -52.02 & -12.60 & 23.60 & N/A \\
			VVC \cite{bross2021overview} & N/A & 46.902 & 0.153 & -19.54 & -56.07 & -13.92 & 19.98 & N/A \\
			\midrule
			\textbf{Learned Generative} & & & & & & & & \\			
			HiFiC \cite{mentzer2020high} & 184.27 & 0.537 & 1.453 & \underline{-72.90} & -65.09 & \underline{-67.18} & -75.54 & \underline{-66.46} \\			
			CDC \cite{yang2023lossy} & 68.66 & 0.114 & 1.307 & -56.74 & \underline{-71.50} & -65.06 & -40.12 & -20.17 \\			
			SCR \cite{lee2022selective} & 11.88 & \textbf{0.029} & \textbf{0.032} & 37.26 & -47.84 & -6.76 & -60.51 & -9.71 \\			
			CTC \cite{jeon2023context} & 399.00 & 0.526 & 0.754 & 108.83 & -49.61 & -20.11 & -82.06 & -32.73 \\			
			Control-GIC \cite{li2024once} & 130.69 & 0.097 & 0.115 & -60.42 & -54.12 & -51.22 & \underline{-85.56} & -43.12 \\			
			\midrule	
			\textbf{HVQ-CGIC (Ours)} & 132.30 & \underline{0.091} & \underline{0.101} & \textbf{-87.30} & \textbf{-71.68} & \textbf{-77.20} & \textbf{-92.22} & \textbf{-67.63} \\
			\bottomrule
		\end{tabular}
	}
	\vspace{-10pt}
\end{table*}

\textbf{Visualizing Coding Efficiency.} As visualized in Figure~\ref{fig:efficiency_bubble}, in VQ-based frameworks, encoding speed is inherently constrained by symbol count; larger codebooks or multi-scale architectures typically lead to higher computational overhead during entropy estimation. This bottleneck is exacerbated when high-order context models are employed (detailed in Appendix~\ref{sec:supp_context}). 

By shifting probability modeling into the continuous embedding space, our HyperVQ module enables the parallel output of probability tables and simultaneous arithmetic coding. Consequently, even when accounting for the entropy coding stage, HyperVQ-equipped models not only achieve a better Rate-Distortion (RD) frontier but also exhibit faster end-to-end inference compared to original backbones using static or sequential coding. 

Further details, including the HVQ-CGIC architectural flowchart (Appendix~\ref{sec:supp_arch}), expanded quantitative comparisons (Appendix~\ref{sec:supp_quantitative}), and CUDA operator design (Appendix~\ref{sec:supp_algo}), are provided in the Supplementary Material.

\section{Limitation, discussion and conclusion}
\label{sec:Conclusion}

Although HyperVQ relies on isotropic covariance for efficiency, theoretically limiting cross-dimensional correlation modeling, its shift to continuous-space probability enables fully differentiable end-to-end RD optimization. This plug-and-play framework provides a foundational entropy-coding component for the VQ family, paving the way for precise and dynamic Rate-Distortion-Perception control in next-generation generative image compression.

\bigskip

{
	\small
	\bibliographystyle{plain}
	\bibliography{citation}

@String(ICIP = {IEEE Int. Conf. Image Process.})

@String(ICIP  = {ICIP})

@article{shannon1959coding,
  title={Coding theorems for a discrete source with a fidelity criterion},
  author={Shannon, Claude E and others},
  journal={IRE Nat. Conv. Rec},
  volume={4},
  number={142-163},
  pages={1},
  year={1959}
}

@misc{bellard2018bpg,
  author = {Fabrice Bellard},
  title = {{BPG} Image format},
  howpublished = {\url{https://bellard.org/bpg/}},
  year = {2018}
}

@article{balle2016end,
  title={End-to-end optimized image compression},
  author={Ball{\'e}, Johannes and Laparra, Valero and Simoncelli, Eero P},
  journal={arXiv preprint arXiv:1611.01704},
  year={2016}
}

@article{balle2018variational,
  title={Variational image compression with a scale hyperprior},
  author={Ball{\'e}, Johannes and Minnen, David and Singh, Saurabh and Hwang, Sung Jin and Johnston, Nick},
  journal={arXiv preprint arXiv:1802.01436},
  year={2018}
}

@article{minnen2018joint,
  title={Joint autoregressive and hierarchical priors for learned image compression},
  author={Minnen, David and Ball{\'e}, Johannes and Toderici, George D},
  journal={Advances in neural information processing systems},
  volume={31},
  year={2018}
}

@inproceedings{minnen2020channel,
  title={Channel-wise autoregressive entropy models for learned image compression},
  author={Minnen, David and Singh, Saurabh},
  booktitle={2020 IEEE International Conference on Image Processing (ICIP)},
  pages={3339--3343},
  year={2020},
  organization={IEEE}
}

@inproceedings{agustsson2019generative,
  title={Generative adversarial networks for extreme learned image compression},
  author={Agustsson, Eirikur and Tschannen, Michael and Mentzer, Fabian and Timofte, Radu and Gool, Luc Van},
  booktitle={Proceedings of the IEEE/CVF international conference on computer vision},
  pages={221--231},
  year={2019}
}

@article{mentzer2020high,
  title={High-fidelity generative image compression},
  author={Mentzer, Fabian and Toderici, George D and Tschannen, Michael and Agustsson, Eirikur},
  journal={Advances in neural information processing systems},
  volume={33},
  pages={11913--11924},
  year={2020}
}

@inproceedings{johnston2018improved,
  title={Improved lossy image compression with priming and spatially adaptive bit rates for recurrent networks},
  author={Johnston, Nick and Vincent, Damien and Minnen, David and Covell, Michele and Singh, Saurabh and Chinen, Troy and Hwang, Sung Jin and Shor, Joel and Toderici, George},
  booktitle={Proceedings of the IEEE conference on computer vision and pattern recognition},
  pages={4385--4393},
  year={2018}
}

@inproceedings{choi2019variable,
  title={Variable rate deep image compression with a conditional autoencoder},
  author={Choi, Yoojin and El-Khamy, Mostafa and Lee, Jungwon},
  booktitle={Proceedings of the IEEE/CVF international conference on computer vision},
  pages={3146--3154},
  year={2019}
}

@inproceedings{yang2021slimmable,
  title={Slimmable compressive autoencoders for practical neural image compression},
  author={Yang, Fei and Herranz, Luis and Cheng, Yongmei and Mozerov, Mikhail G},
  booktitle={Proceedings of the IEEE/CVF Conference on Computer Vision and Pattern Recognition},
  pages={4998--5007},
  year={2021}
}

@inproceedings{iwai2024controlling,
  title={Controlling rate, distortion, and realism: Towards a single comprehensive neural image compression model},
  author={Iwai, Shoma and Miyazaki, Tomo and Omachi, Shinichiro},
  booktitle={Proceedings of the IEEE/CVF Winter Conference on Applications of Computer Vision},
  pages={2900--2909},
  year={2024}
}

@inproceedings{esser2021taming,
  title={Taming transformers for high-resolution image synthesis},
  author={Esser, Patrick and Rombach, Robin and Ommer, Bjorn},
  booktitle={Proceedings of the IEEE/CVF conference on computer vision and pattern recognition},
  pages={12873--12883},
  year={2021}
}

@article{yang2023lossy,
  title={Lossy image compression with conditional diffusion models},
  author={Yang, Ruihan and Mandt, Stephan},
  journal={Advances in Neural Information Processing Systems},
  volume={36},
  pages={64971--64995},
  year={2023}
}

@inproceedings{jia2024generative,
  title={Generative latent coding for ultra-low bitrate image compression},
  author={Jia, Zhaoyang and Li, Jiahao and Li, Bin and Li, Houqiang and Lu, Yan},
  booktitle={Proceedings of the IEEE/CVF Conference on Computer Vision and Pattern Recognition},
  pages={26088--26098},
  year={2024}
}

@article{lee2022selective,
  title={Selective compression learning of latent representations for variable-rate image compression},
  author={Lee, Jooyoung and Jeong, Seyoon and Kim, Munchurl},
  journal={Advances in Neural Information Processing Systems},
  volume={35},
  pages={13146--13157},
  year={2022}
}

@article{li2024once,
  title={Once-for-all: Controllable generative image compression with dynamic granularity adaption},
  author={Li, Anqi and Li, Feng and Liu, Yuxi and Cong, Runmin and Zhao, Yao and Bai, Huihui},
  journal={arXiv preprint arXiv:2406.00758},
  year={2024}
}

@article{bross2021overview,
  title={Overview of the versatile video coding (VVC) standard and its applications},
  author={Bross, Benjamin and Wang, Ye-Kui and Ye, Yan and Liu, Shan and Chen, Jianle and Sullivan, Gary J and Ohm, Jens-Rainer},
  journal={IEEE Transactions on Circuits and Systems for Video Technology},
  volume={31},
  number={10},
  pages={3736--3764},
  year={2021},
  publisher={IEEE}
}

@inproceedings{muckley2023improving,
  title={Improving statistical fidelity for neural image compression with implicit local likelihood models},
  author={Muckley, Matthew J and El-Nouby, Alaaeldin and Ullrich, Karen and J{\'e}gou, Herv{\'e} and Verbeek, Jakob},
  booktitle={International Conference on Machine Learning},
  pages={25426--25443},
  year={2023},
  organization={PMLR}
}

@inproceedings{jeon2023context,
  title={Context-based trit-plane coding for progressive image compression},
  author={Jeon, Seungmin and Choi, Kwang Pyo and Park, Youngo and Kim, Chang-Su},
  booktitle={Proceedings of the IEEE/CVF Conference on Computer Vision and Pattern Recognition},
  pages={14348--14357},
  year={2023}
}

@misc{kodak_dataset,
  title = {Kodak Lossless True Color Image Suite},
  author = {Eastman Kodak Company},
  howpublished = {\url{http://r0k.us/graphics/kodak/}},
  note = {Original images, commonly used as a test set for image compression.},
  year = {1993}
}

@inproceedings{agustsson2017ntire,
  title={Ntire 2017 challenge on single image super-resolution: Dataset and study},
  author={Agustsson, Eirikur and Timofte, Radu},
  booktitle={Proceedings of the IEEE conference on computer vision and pattern recognition workshops},
  pages={126--135},
  year={2017}
}

@article{toderici2020clic,
  title={Clic 2020: Challenge on learned image compression},
  author={Toderici, George and Theis, Lucas and Johnston, Nick and Agustsson, Eirikur and Mentzer, Fabian and Ball{\'e}, Johannes and Shi, Wenzhe and Timofte, Radu},
  journal={Retrieved March},
  volume={29},
  pages={2021},
  year={2020}
}

@inproceedings{rippel2017real,
  title={Real-time adaptive image compression},
  author={Rippel, Oren and Bourdev, Lubomir},
  booktitle={International conference on machine learning},
  pages={2922--2930},
  year={2017},
  organization={PMLR}
}

@inproceedings{li2018learning,
  title={Learning convolutional networks for content-weighted image compression},
  author={Li, Mu and Zuo, Wangmeng and Gu, Shuhang and Zhao, Debin and Zhang, David},
  booktitle={Proceedings of the IEEE conference on computer vision and pattern recognition},
  pages={3214--3223},
  year={2018}
}

@inproceedings{shin2022expanded,
  title={Expanded adaptive scaling normalization for end to end image compression},
  author={Shin, Chajin and Lee, Hyeongmin and Son, Hanbin and Lee, Sangjin and Lee, Dogyoon and Lee, Sangyoun},
  booktitle={European Conference on Computer Vision},
  pages={390--405},
  year={2022},
  organization={Springer}
}

@inproceedings{pan2022content,
  title={Content adaptive latents and decoder for neural image compression},
  author={Pan, Guanbo and Lu, Guo and Hu, Zhihao and Xu, Dong},
  booktitle={European Conference on Computer Vision},
  pages={556--573},
  year={2022},
  organization={Springer}
}

@inproceedings{li2022content,
  title={Content-oriented learned image compression},
  author={Li, Meng and Gao, Shangyin and Feng, Yihui and Shi, Yibo and Wang, Jing},
  booktitle={European Conference on Computer Vision},
  pages={632--647},
  year={2022},
  organization={Springer}
}

@inproceedings{zhu2022transformer,
  title={Transformer-based transform coding},
  author={Zhu, Yinhao and Yang, Yang and Cohen, Taco},
  booktitle={International conference on learning representations},
  year={2022}
}

@inproceedings{liu2023learned,
  title={Learned image compression with mixed transformer-cnn architectures},
  author={Liu, Jinming and Sun, Heming and Katto, Jiro},
  booktitle={Proceedings of the IEEE/CVF conference on computer vision and pattern recognition},
  pages={14388--14397},
  year={2023}
}

@inproceedings{careil2023towards,
  title={Towards image compression with perfect realism at ultra-low bitrates},
  author={Careil, Marlene and Muckley, Matthew J and Verbeek, Jakob and Lathuili{\`e}re, St{\'e}phane},
  booktitle={The Twelfth International Conference on Learning Representations},
  year={2023}
}

@inproceedings{mentzer2018conditional,
  title={Conditional probability models for deep image compression},
  author={Mentzer, Fabian and Agustsson, Eirikur and Tschannen, Michael and Timofte, Radu and Van Gool, Luc},
  booktitle={Proceedings of the IEEE conference on computer vision and pattern recognition},
  pages={4394--4402},
  year={2018}
}

@article{lee2018context,
  title={Context-adaptive entropy model for end-to-end optimized image compression},
  author={Lee, Jooyoung and Cho, Seunghyun and Beack, Seung-Kwon},
  journal={arXiv preprint arXiv:1809.10452},
  year={2018}
}

@inproceedings{cheng2020learned,
  title={Learned image compression with discretized gaussian mixture likelihoods and attention modules},
  author={Cheng, Zhengxue and Sun, Heming and Takeuchi, Masaru and Katto, Jiro},
  booktitle={Proceedings of the IEEE/CVF conference on computer vision and pattern recognition},
  pages={7939--7948},
  year={2020}
}

@inproceedings{lin2020spatial,
  title={A spatial rnn codec for end-to-end image compression},
  author={Lin, Chaoyi and Yao, Jiabao and Chen, Fangdong and Wang, Li},
  booktitle={Proceedings of the IEEE/CVF Conference on Computer Vision and Pattern Recognition},
  pages={13269--13277},
  year={2020}
}

@inproceedings{zhang2021attention,
  title={Attention-guided image compression by deep reconstruction of compressive sensed saliency skeleton},
  author={Zhang, Xi and Wu, Xiaolin},
  booktitle={Proceedings of the IEEE/CVF Conference on Computer Vision and Pattern Recognition},
  pages={13354--13364},
  year={2021}
}

@inproceedings{he2021checkerboard,
  title={Checkerboard context model for efficient learned image compression},
  author={He, Dailan and Zheng, Yaoyan and Sun, Baocheng and Wang, Yan and Qin, Hongwei},
  booktitle={Proceedings of the IEEE/CVF Conference on Computer Vision and Pattern Recognition},
  pages={14771--14780},
  year={2021}
}

@inproceedings{gao2021neural,
  title={Neural image compression via attentional multi-scale back projection and frequency decomposition},
  author={Gao, Ge and You, Pei and Pan, Rong and Han, Shunyuan and Zhang, Yuanyuan and Dai, Yuchao and Lee, Hojae},
  booktitle={Proceedings of the IEEE/CVF International Conference on Computer Vision},
  pages={14677--14686},
  year={2021}
}

@inproceedings{kim2022joint,
  title={Joint global and local hierarchical priors for learned image compression},
  author={Kim, Jun-Hyuk and Heo, Byeongho and Lee, Jong-Seok},
  booktitle={Proceedings of the IEEE/CVF Conference on Computer Vision and Pattern Recognition},
  pages={5992--6001},
  year={2022}
}

@article{zhang2022multi,
  title={Multi-modality deep restoration of extremely compressed face videos},
  author={Zhang, Xi and Wu, Xiaolin},
  journal={IEEE Transactions on Pattern Analysis and Machine Intelligence},
  volume={45},
  number={2},
  pages={2024--2037},
  year={2022},
  publisher={IEEE}
}

@inproceedings{he2022elic,
  title={Elic: Efficient learned image compression with unevenly grouped space-channel contextual adaptive coding},
  author={He, Dailan and Yang, Ziming and Peng, Weikun and Ma, Rui and Qin, Hongwei and Wang, Yan},
  booktitle={Proceedings of the IEEE/CVF conference on computer vision and pattern recognition},
  pages={5718--5727},
  year={2022}
}

@inproceedings{jiang2023mlic,
  title={Mlic: Multi-reference entropy model for learned image compression},
  author={Jiang, Wei and Yang, Jiayu and Zhai, Yongqi and Ning, Peirong and Gao, Feng and Wang, Ronggang},
  booktitle={Proceedings of the 31st ACM International Conference on Multimedia},
  pages={7618--7627},
  year={2023}
}

@article{wang2023evc,
  title={Evc: Towards real-time neural image compression with mask decay},
  author={Wang, Guo-Hua and Li, Jiahao and Li, Bin and Lu, Yan},
  journal={arXiv preprint arXiv:2302.05071},
  year={2023}
}

@article{agustsson2017soft,
  title={Soft-to-hard vector quantization for end-to-end learning compressible representations},
  author={Agustsson, Eirikur and Mentzer, Fabian and Tschannen, Michael and Cavigelli, Lukas and Timofte, Radu and Benini, Luca and Gool, Luc V},
  journal={Advances in neural information processing systems},
  volume={30},
  year={2017}
}

@inproceedings{zhang2023lvqac,
  title={Lvqac: Lattice vector quantization coupled with spatially adaptive companding for efficient learned image compression},
  author={Zhang, Xi and Wu, Xiaolin},
  booktitle={Proceedings of the IEEE/CVF Conference on Computer Vision and Pattern Recognition},
  pages={10239--10248},
  year={2023}
}

@inproceedings{feng2023nvtc,
  title={Nvtc: Nonlinear vector transform coding},
  author={Feng, Runsen and Guo, Zongyu and Li, Weiping and Chen, Zhibo},
  booktitle={Proceedings of the IEEE/CVF Conference on Computer Vision and Pattern Recognition},
  pages={6101--6110},
  year={2023}
}

@article{lei2024approaching,
  title={Approaching rate-distortion limits in neural compression with lattice transform coding},
  author={Lei, Eric and Hassani, Hamed and Bidokhti, Shirin Saeedi},
  journal={arXiv preprint arXiv:2403.07320},
  year={2024}
}

@article{zhang2024learning,
  title={Learning optimal lattice vector quantizers for end-to-end neural image compression},
  author={Zhang, Xi and Wu, Xiaolin},
  journal={Advances in Neural Information Processing Systems},
  volume={37},
  pages={106497--106518},
  year={2024}
}

@article{russakovsky2015imagenet,
  title={Imagenet large scale visual recognition challenge},
  author={Russakovsky, Olga and Deng, Jia and Su, Hao and Krause, Jonathan and Satheesh, Sanjeev and Ma, Sean and Huang, Zhiheng and Karpathy, Andrej and Khosla, Aditya and Bernstein, Michael and others},
  journal={International journal of computer vision},
  volume={115},
  number={3},
  pages={211--252},
  year={2015},
  publisher={Springer}
}

@inproceedings{zhu2022unified,
  title={Unified multivariate gaussian mixture for efficient neural image compression},
  author={Zhu, Xiaosu and Song, Jingkuan and Gao, Lianli and Zheng, Feng and Shen, Heng Tao},
  booktitle={Proceedings of the IEEE/CVF Conference on Computer Vision and Pattern Recognition},
  pages={17612--17621},
  year={2022}
}

@inproceedings{mao2024extreme,
  title={Extreme image compression using fine-tuned vqgans},
  author={Mao, Qi and Yang, Tinghan and Zhang, Yinuo and Wang, Zijian and Wang, Meng and Wang, Shiqi and Jin, Libiao and Ma, Siwei},
  booktitle={2024 Data Compression Conference (DCC)},
  pages={203--212},
  year={2024},
  organization={IEEE}
}

@article{zheng2022movq,
  title={Movq: Modulating quantized vectors for high-fidelity image generation},
  author={Zheng, Chuanxia and Vuong, Tung-Long and Cai, Jianfei and Phung, Dinh},
  journal={Advances in Neural Information Processing Systems},
  volume={35},
  pages={23412--23425},
  year={2022}
}

@article{zhu2024scaling,
  title={Scaling the codebook size of vq-gan to 100,000 with a utilization rate of 99\%},
  author={Zhu, Lei and Wei, Fangyun and Lu, Yanye and Chen, Dong},
  journal={Advances in Neural Information Processing Systems},
  volume={37},
  pages={12612--12635},
  year={2024}
}

@inproceedings{korber2024egic,
  title={Egic: enhanced low-bit-rate generative image compression guided by semantic segmentation},
  author={K{\"o}rber, Nikolai and Kromer, Eduard and Siebert, Andreas and Hauke, Sascha and Mueller-Gritschneder, Daniel and Schuller, Bj{\"o}rn},
  booktitle={European Conference on Computer Vision},
  pages={202--220},
  year={2024},
  organization={Springer}
}
}

\newpage
\setcounter{page}{1}

\setcounter{section}{0}
\setcounter{equation}{0}
\renewcommand\thesection{\Alph{section}}
\renewcommand\thesubsection{\thesection.\arabic{subsection}}
\renewcommand\theequation{S\arabic{equation}}
\renewcommand\thefigure{S\arabic{figure}}
\renewcommand\thetable{S\arabic{table}}

\begin{center}
	\Large \textbf{Supplementary Material}
\end{center}

This supplementary material serves to substantiate the findings presented in the main paper, offering deep theoretical analysis, detailed architecture design, granular ablation data, and extended visual comparisons. The content is organized as follows:

\begin{itemize}
	\item \textbf{Appendix \ref{sec:supp_quantitative}: Extended Quantitative RD Comparisons.} Provides extended Rate-Distortion curves, including comparisons against non-VQ generative methods and evaluations on high-resolution benchmarks (CLIC 2020, DIV2K).
	\item \textbf{Appendix \ref{sec:supp_algo}: CUDA-Accelerated Entropy Preparation (PEE).} Provides the mathematical definition and pseudo-code for the GPU-based Cumulative Distribution Function (CDF) construction, along with our bounded tail-floor approximation for ultra-large codebooks.
	\item \textbf{Appendix \ref{sec:supp_training}: Integration Protocol.} Details a rapid adaptation method to equip any VQ backbone with HyperVQ, alongside an ablation study justifying our joint-training recommendation.
	\item \textbf{Appendix \ref{sec:supp_theory}: Theoretical Derivation.} Elucidates the complete mathematical derivation from the rate-distortion objective to the differentiable loss function, including a formal justification for the centroid approximation.
	\item \textbf{Appendix \ref{sec:supp_arch}: Detailed Architecture of HVQ-CGIC.} Presents the complete flowchart of the instantiation on the controllable multi-granularity codec.
	\item \textbf{Appendix \ref{sec:supp_context}: Comparison with Traditional High-Order Context Modeling.} Evaluates HyperVQ against manually designed spatial context arithmetic coders in terms of compression bounds and inference latency.
	\item \textbf{Appendix \ref{sec:supp_kway}: Analysis: Why Not a Direct K-Way Classifier?} Details why our continuous embedding-space Gaussian modeling is vastly superior to predicting probabilities via a K-way classifier.
	\item \textbf{Appendix \ref{sec:supp_entropy_bit}: Entropy Coding and Spatial Bit Usage Analysis.} Visually demonstrates how HVQ-CGIC exhausts coding redundancy in sparse latent spaces.
	\item \textbf{Appendix \ref{sec:supp_visuals}: Extended Qualitative Visual Comparisons.} Offers extensive high-resolution visual comparisons across multiple datasets to demonstrate the superior perceptual fidelity of our framework.
\end{itemize}

\section{Extended Quantitative RD Comparisons}
\label{sec:supp_quantitative}

In this section, we present two sets of extended quantitative comparisons to further validate the superiority of the proposed framework. First, we benchmark HVQ-CGIC against recent state-of-the-art \textit{non-VQ} continuous-latent generative codecs. Second, we evaluate the method's high-resolution generalization capability on the CLIC 2020 and DIV2K datasets.

\subsection{Comparison with Non-VQ Generative Codecs}
\label{sec:supp_non_vq_comp} 
\begin{figure}[htbp]
	\centering
	\includegraphics[width=0.8\linewidth]{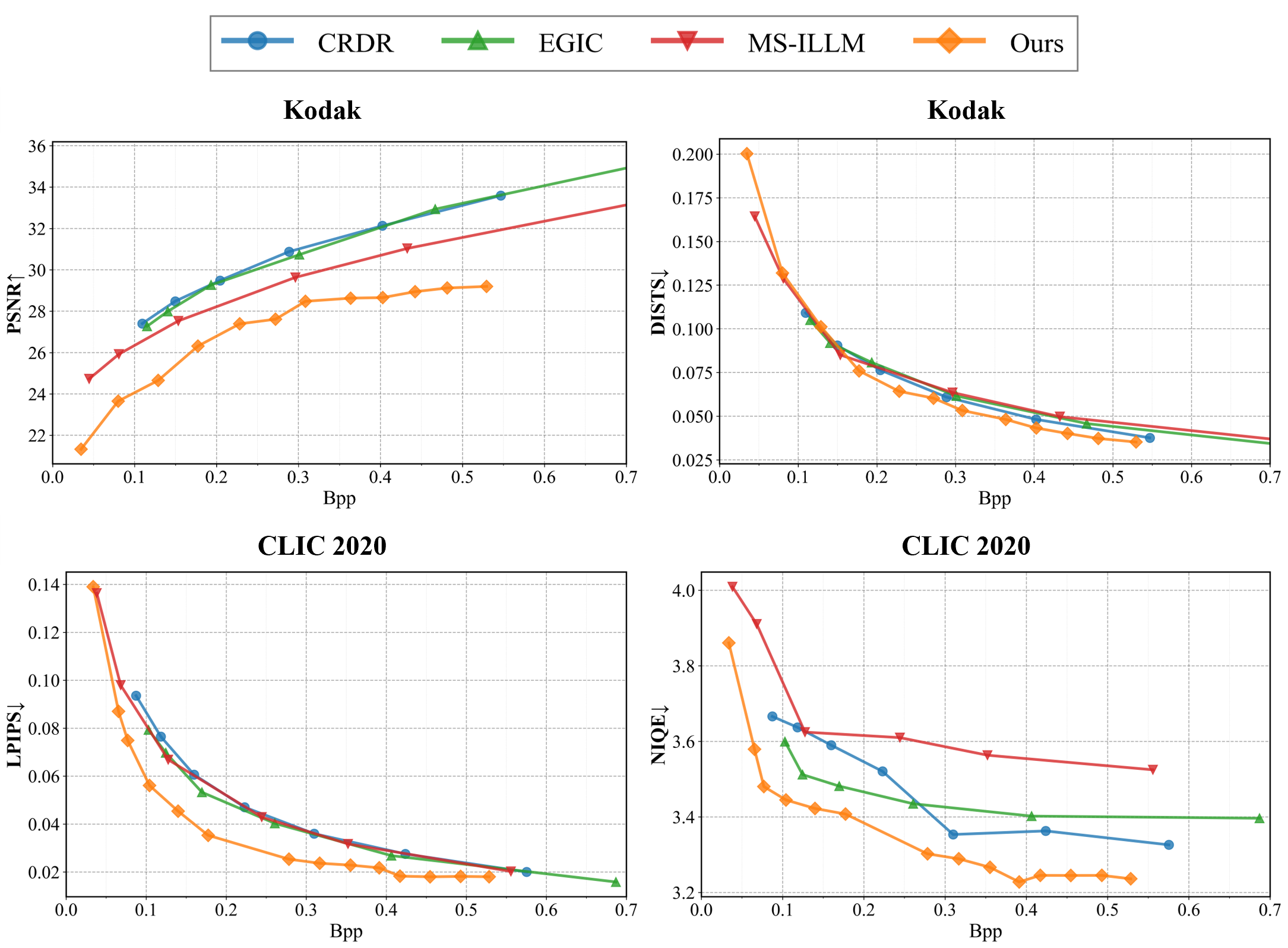}
	\caption{\textbf{Comparison with non-VQ generative methods} on Kodak and CLIC 2020 test sets.}
	\label{fig:supp_non_vq}
\end{figure}

To provide a comprehensive evaluation across different generative paradigms, we compare HVQ-CGIC against MS-ILLM\cite{muckley2023improving}, CRDR\cite{iwai2024controlling}, and EGIC\cite{korber2024egic}. As shown in Figure~\ref{fig:supp_non_vq}, HVQ-CGIC establishes a highly competitive Pareto frontier against these advanced continuous-latent models. This strongly validates the potential of VQ-based compression when equipped with powerful, content-adaptive hyperprior entropy modeling.

\subsection{Performance on High-Resolution Benchmarks}
\begin{figure*}[htbp]
	\centering
	\includegraphics[width=1.0\linewidth]{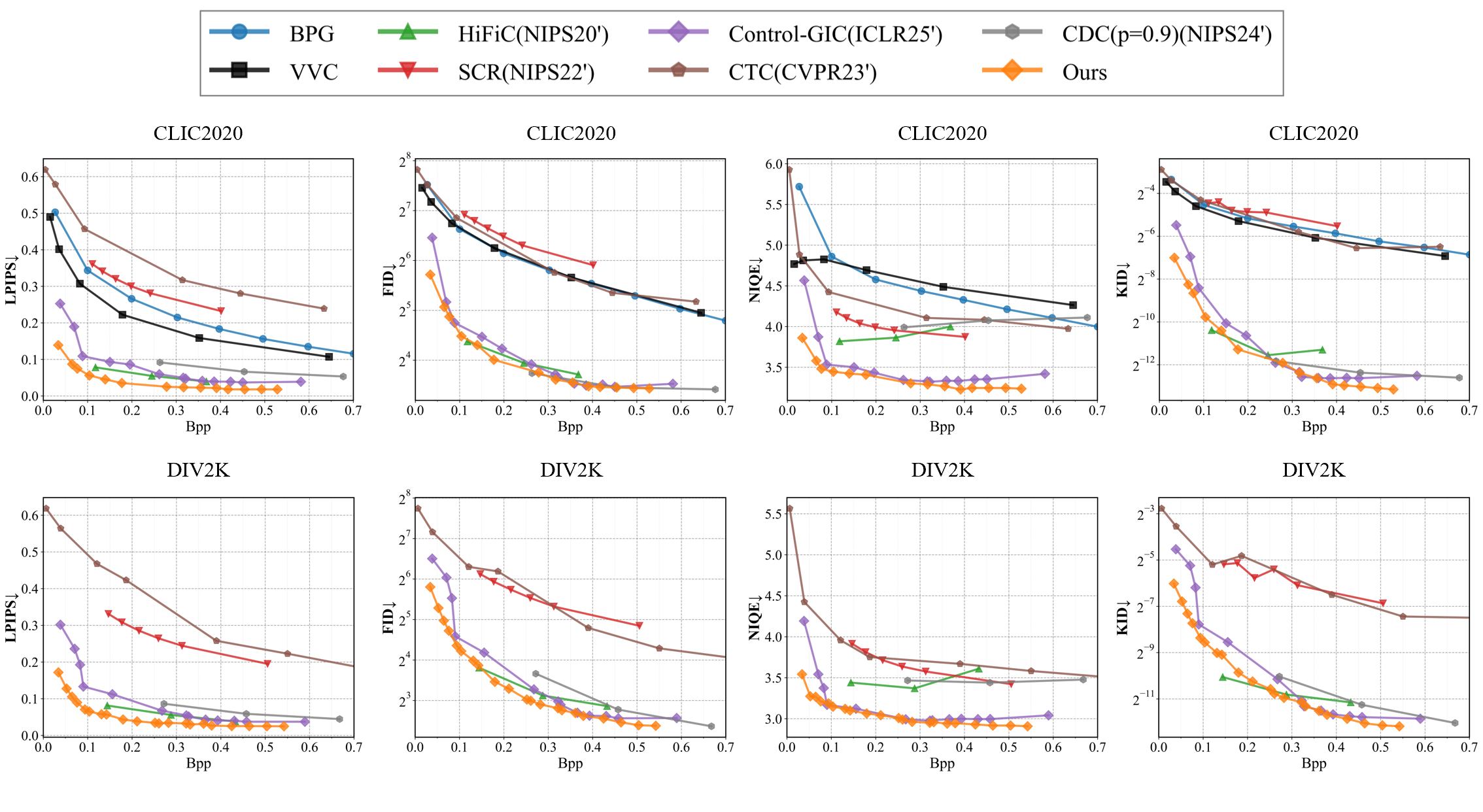}
	\caption{\textbf{High-Resolution Generalization.} Comparison of methods on the CLIC 2020 professional set (top) and DIV2K (bottom).}
	\label{fig:supp_clic_rd}
\end{figure*}

We evaluate the generalization capability of HVQ-CGIC on high-resolution benchmarks: the CLIC 2020 Professional Set \cite{toderici2020clic} and DIV2K \cite{agustsson2017ntire}. The Rate-Distortion curves are presented in Figure~\ref{fig:supp_clic_rd}. 

\textbf{Performance on CLIC 2020.} As shown in the top row of Figure~\ref{fig:supp_clic_rd}, HVQ-CGIC maintains significant advantages in perceptual metrics (LPIPS, FID, KID) on the professional photography dataset. This confirms that our entropy-based routing strategy generalizes well to complex lighting and composition without requiring dataset-specific retraining.

\textbf{Performance on DIV2K.} On the 2K-resolution DIV2K dataset (Figure~\ref{fig:supp_clic_rd}, bottom row), HVQ-CGIC consistently outperforms the baselines across the measured metrics. This demonstrates the robustness of our hyperprior modeling in handling rich high-frequency details at scaling resolutions.

\section{CUDA-Accelerated Entropy Preparation}
\label{sec:supp_algo}

To reduce the entropy coding latency in high-resolution image compression, we migrate the Cumulative Distribution Function (CDF) preparation process from the CPU to the GPU by designing a custom CUDA-accelerated entropy preparation operator. This operator does not replace the final arithmetic coder; rather, it efficiently transforms the network-output probability parameters into the integer CDFs, lower bounds, and frequencies directly consumable by the arithmetic coder.

\subsection{Categorical CDF Preparation for Index Stream}
Let $\mathbf{l}_{n} \in \mathbb{R}^{K}$ denote the logits at symbol position $n$, where $K$ is the alphabet size. We first convert logits into probabilities:
\begin{equation}
	\pi_{n,k} = \frac{\exp(l_{n,k})}{\sum_{j=0}^{K-1}\exp(l_{n,j})}, \quad k=0, \dots,K-1.
\end{equation}
Given an entropy precision of $p$ bits, the integer scale is $S = 2^{p}$. To guarantee a valid integer CDF with a minimum increment of one for each symbol, we use monotonic quantization:
\begin{equation}
	\hat{C}_{n,0} = 0, \quad \hat{C}_{n,K} = S,
\end{equation}
and for $k=1,\dots,K-1$,
\begin{equation}
	\hat{C}_{n,k} = \min\!\left( \left\lfloor (S-K)\sum_{j=0}^{k-1}\pi_{n,j}\right\rfloor + k,\; S-(K-k) \right).
\end{equation}
For the target symbol index $i_n \in \{0,\dots,K-1\}$, the arithmetic encoder consumes the lower bound $L^{(idx)}_n$ and frequency $F^{(idx)}_n$:
\begin{equation}
	L^{(idx)}_n = \hat{C}_{n,i_n}, \quad F^{(idx)}_n = \hat{C}_{n,i_n+1} - \hat{C}_{n,i_n}.
\end{equation}

\subsection{Gaussian CDF Preparation for Hyper-latents}
For the hyper-latent stream, we discretize the Gaussian distribution on the support $v \in [v_{\min}, v_{\max}] \cap \mathbb{Z}$, and compute the unnormalized mass by $q_c(v) = \exp\!\left(-\frac{v^2}{2\sigma_c^2}\right)$. After normalization:
\begin{equation}
	p_c(v) = \frac{q_c(v)}{\sum_{u=v_{\min}}^{v_{\max}} q_c(u)}.
\end{equation}
The corresponding integer CDF $\tilde{C}_{c,:}$ is constructed using the same monotonic quantization rule. For symbol $z_{c,m}$, let $r_{c,m} = z_{c,m} - v_{\min}$. The arithmetic encoder consumes the corresponding lower bound $L^{(hyp)}_{c,m}$ and frequency $F^{(hyp)}_{c,m}$:
\begin{equation}
	L^{(hyp)}_{c,m} = \tilde{C}_{c,r_{c,m}}, \quad F^{(hyp)}_{c,m} = \tilde{C}_{c,r_{c,m}+1} - \tilde{C}_{c,r_{c,m}}.
\end{equation}

\begin{algorithm}[t]
	\caption{CUDA-Accelerated Entropy Preparation}
	\label{alg:cuda_entropy_prep}
	\begin{algorithmic}[1]
		\Require Categorical logits $\mathbf{l}\in\mathbb{R}^{N\times K}$, target indices $\mathbf{i}\in\{0,\dots,K-1\}^{N}$; hyper-latent symbols $\mathbf{z}$, channel-wise scales $\boldsymbol{\sigma}$, precision $p$.
		\Ensure Integer lower bounds $\mathbf{L}^{(idx)}, \mathbf{L}^{(hyp)}$, frequencies $\mathbf{F}^{(idx)}, \mathbf{F}^{(hyp)}$, and CDFs $\hat{\mathbf{C}}, \tilde{\mathbf{C}}$.
		
		\State Set integer scale $S \gets 2^p$
		
		\Statex \textbf{// Stage 1: Categorical CDF Preparation (Index Stream)}
		\For{each spatial position $n \in \{1, \dots, N\}$ \textbf{in parallel}}
		\State $\pi_{n,k} \gets \frac{\exp(l_{n,k})}{\sum_j \exp(l_{n,j})}$ for $k=0,\dots,K-1$ \Comment{Softmax}
		\State Compute integer CDF $\hat{C}_{n,:}$ via monotonic quantization to scale $S$
		\State $L^{(idx)}_n \gets \hat{C}_{n,i_n}$
		\State $F^{(idx)}_n \gets \hat{C}_{n,i_n+1} - \hat{C}_{n,i_n}$
		\EndFor
		
		\Statex \textbf{// Stage 2: Gaussian CDF Preparation (Hyper-latent Stream)}
		\For{each channel $c$ \textbf{in parallel}}
		\State $q_c(v) \gets \exp(-v^2 / 2\sigma_c^2)$ for $v \in [v_{\min}, v_{\max}] \cap \mathbb{Z}$
		\State $p_c(v) \gets q_c(v) / \sum_u q_c(u)$ \Comment{Normalize PMF}
		\State Quantize $p_c(v)$ into integer CDF $\tilde{C}_{c,:}$ using monotonic rule
		\EndFor
		
		\For{each hyper-latent symbol $z_{c,m}$ \textbf{in parallel}}
		\State $r_{c,m} \gets z_{c,m} - v_{\min}$ \Comment{Calculate table offset}
		\State $L^{(hyp)}_{c,m} \gets \tilde{C}_{c,r_{c,m}}$
		\State $F^{(hyp)}_{c,m} \gets \tilde{C}_{c,r_{c,m}+1} - \tilde{C}_{c,r_{c,m}}$
		\EndFor
		
		\State \Return $\mathbf{L}^{(idx)}, \mathbf{F}^{(idx)}, \hat{\mathbf{C}}$ and $\mathbf{L}^{(hyp)}, \mathbf{F}^{(hyp)}, \tilde{\mathbf{C}}$
	\end{algorithmic}
\end{algorithm}

\subsection{Bounded Tail-Floor for Ultra-Large Codebooks}
\label{sec:supp_tail_floor}
For standard codebooks ($K \le 8192$), evaluating the exact Euclidean distance to all anchors is highly parallelizable and practically instantaneous. However, for ultra-large codebooks (e.g., VQGAN-LC with $K=100,000$), calculating a full exact Softmax introduces unnecessary memory and latency overheads. 

To resolve this, our PEE utilizes a \textbf{Bounded Tail-Floor Approximation}. Since our logits are strictly monotonically decreasing with geometric distance (Equation \ref{eq:hypervq_prob}), anchors positioned far from the predicted mean $\boldsymbol{\mu}_{b,ij}$ contribute almost zero mass to the true categorical distribution. Therefore, during the CDF construction phase in the CUDA kernel, we evaluate exact distances only for a set of candidate anchors within a threshold radius. For all remote, "pruned" anchors, we bypass the exponentiation step and directly assign them the minimum valid integer frequency ($\hat{F}_{k} = 1$) in the arithmetic integer scale $S$. 

This strategy ensures three critical advantages: 1) It maintains a strictly legal and monotonic CDF, guaranteeing 100\% mathematical decodability without any bitstream format changes; 2) The decoder replicates the exact same deterministic tail-floor rule based on the decoded $\boldsymbol{\mu}$; 3) It slashes the large-codebook entropy preparation latency to a mere $\sim 10$ms penalty. We emphasize that the compression performance reported in Table 1 of the main paper is obtained using this Bounded Tail-Floor Approximation within our PEE. As shown in Table \ref{table:tail_floor_comp}, the gap between this efficient bounded estimation and the exact dense Softmax is less than 0.2\% in bitrate gain, justifying the use of the bounded version as our default practical configuration.

\begin{table}[htbp]
	\centering
	\caption{Ablation Study of Bounded Tail-Floor Approximation on VQGAN-LC (ImageNet-1K Val). The results show that the bitrate gap between the proposed approximation and the exact dense Softmax is negligible (within 0.2\% gain), justifying our efficient tail-floor design.}
	\label{table:tail_floor_comp}
	\resizebox{0.9\linewidth}{!}{
		\begin{tabular}{lccccc}
			\toprule
			\textbf{Method} & \textbf{Codebook ($K$)} & \textbf{Scale} & \textbf{Theo. bpp} & \textbf{HyperVQ (Approx.)} & \textbf{HyperVQ (Exact Softmax)} \\
			\midrule
			VQGAN-LC & 16384 & $16\times$ & 0.0547 & 0.0458 (\textcolor{teal}{16.3\%}) & 0.0458 (\textcolor{gaincolor}{16.3\%}) \\
			VQGAN-LC & 100K  & $16\times$ & 0.0641 & 0.0564 (\textcolor{teal}{12.0\%}) & 0.0562 (\textcolor{gaincolor}{12.5\%}) \\
			VQGAN-LC & 100K  & $8\times$  & 0.2595 & 0.2072 (\textcolor{teal}{20.2\%}) & 0.2071 (\textcolor{gaincolor}{20.2\%}) \\
			\bottomrule
		\end{tabular}
	}
\end{table}

\subsection{Acceleration Analysis and Timing Disambiguation}
It is important to clarify the relationship between the system-level timings reported in the subsequent tables (e.g., Table~\ref{tab:context_ablation}) and the kernel-level efficiency of our proposed module. 

The ultra-low encoding and decoding latencies (e.g., $2.71$ ms and $4.17$ ms) presented in the context ablation study represent the per-image throughput measured under a \textbf{batched parallel inference} setting on the Kodak dataset, which fully saturates the GPU resources and demonstrates the theoretical peak pipeline speed. 

In contrast, the micro-benchmark analysis below isolates the computational acceleration of the custom CDF preparation kernel itself on a \textbf{single, unbatched} high-resolution ($1920\times 1080$) image. In this kernel-level test, we measure exclusively the CDF preparation cost, deliberately excluding neural-network forward passes, process creation, and file I/O overheads. By offloading this high-dimensional density-to-probability mapping to the GPU, our custom CUDA implementation achieves a massive \textbf{39.8$\times$ speedup} over the baseline sequential CPU execution. In terms of processing throughput, this corresponds to a dramatic increase from $4.60$ MPix/s to $183.0$ MPix/s. 

Furthermore, at the system-prototype level, replacing the original Huffman coding bottleneck with our GPU-accelerated CDF preparation improves the entire end-to-end entropy pipeline by $1.93\times$ for encoding and $2.13\times$ for decoding over a serial CPU pipeline. This validates that our complex content-adaptive entropy modeling introduces negligible latency overhead while unlocking substantial Rate-Distortion gains.

\section{Integration Protocol: How to Apply HyperVQ to Existing VQ Codecs}
\label{sec:supp_training}

Because HyperVQ operates exclusively in the continuous embedding space, it can be rapidly deployed to enhance the entropy modeling of almost any existing VQ-based architecture. In this section, we provide a practical, three-stage progressive integration recipe. 

It is important to note that while \textbf{Stage B} (Rapid Adaptation) provides a highly effective and quick method to slash the bitrate of a pre-trained VQ model, we strongly recommend completing \textbf{Stage C} (Joint End-to-End Training) to achieve the optimal Rate-Distortion balance. We use the HVQ-CGIC instantiation as a case study to ablate and validate this strategy.

\noindent \textbf{Stage A: Baseline VQ Training (or Loading Pre-trained Weights).} \\
We first obtain a high-quality VQ backbone consisting of an encoder, decoder, vector quantizer, and discriminator. Only the distortion and commitment terms are optimized: $\mathcal{L}_{\text{A}} = D(x,\hat{x})$. At this stage, the hyperprior networks are absent, establishing a strong perceptually-aligned reconstruction baseline. (For existing models, one can simply load the pre-trained weights).

\noindent \textbf{Stage B: Rapid Hyperprior Adaptation (Quick Application).} \\
We freeze all the parameters of the base VQ codec. The HyperVQ networks (Analysis and Synthesis transforms) are randomly initialized and trained \textit{exclusively} by minimizing the differentiable rate term: $\mathcal{L}_{\text{B}} = R$. This forces the hyperprior to rapidly learn efficient, content-adaptive probability predictions for the fixed VQ indices produced by the frozen encoder. 

\noindent \textbf{Stage C: Joint RD Fine-tuning (Recommended).} \\
Finally, we unfreeze the entire framework and fine-tune it end-to-end with a reduced learning rate using the full Rate-Distortion objective: $\mathcal{L}_{\text{C}} = D(x,\hat{x}) + R$. This allows the VQ encoder's feature representations and the hyperprior's probability models to co-adapt, actively restructuring the latent space for compressibility.

\begin{figure}[htbp]
	\centering
	\includegraphics[width=0.8\linewidth]{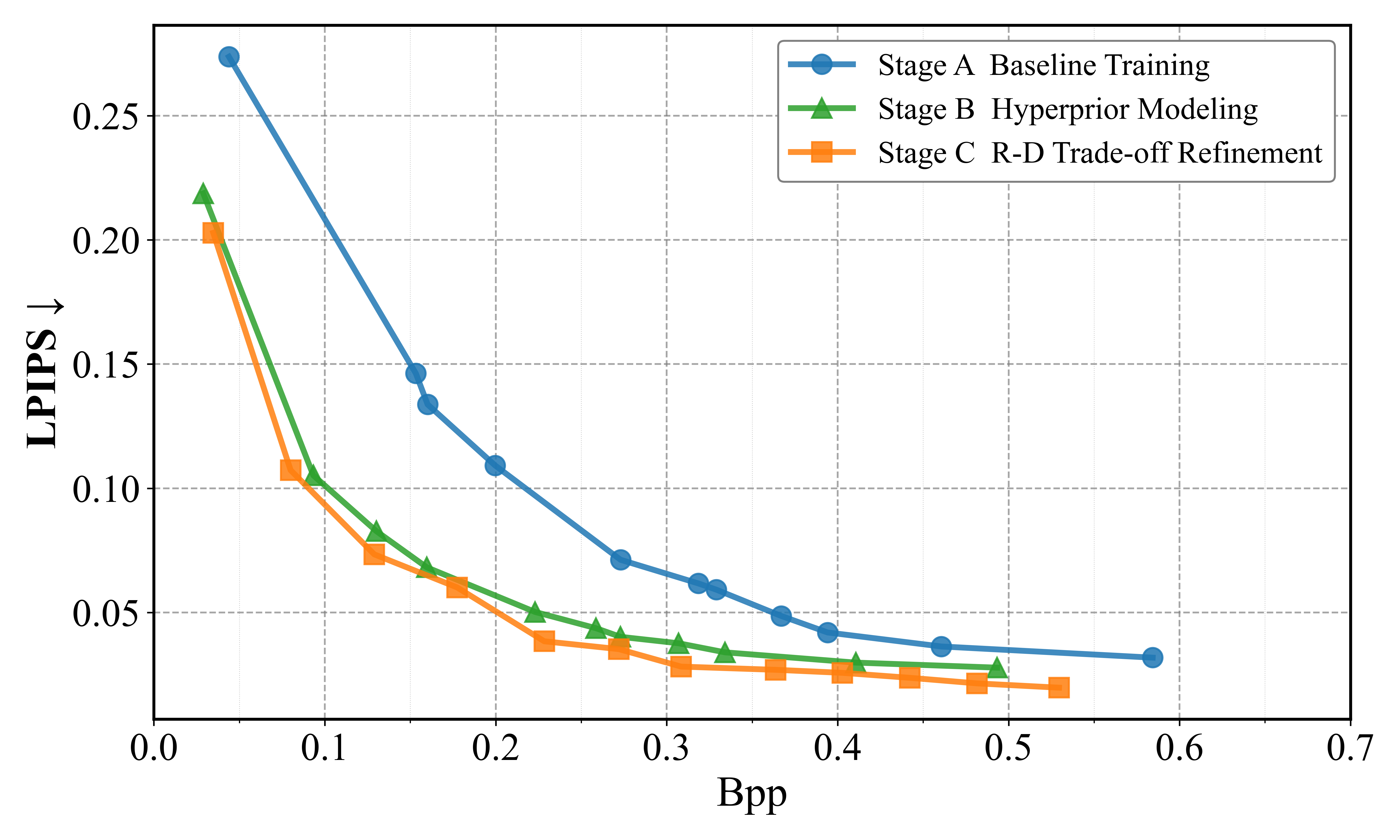}
	\caption{\textbf{Ablation study of the three training stages on Kodak.} Stage B (Rapid Adaptation) significantly drops the bitrate with minimal quality compromise, while Stage C (Joint Fine-tuning) pushes the Pareto frontier further to the lower-left, optimizing the ultimate RD trade-off.}
	\label{fig:supp_ablation}
\end{figure}

\textbf{Ablation Analysis.} The effectiveness of this protocol is validated in Figure~\ref{fig:supp_ablation}. The transition from Stage A to Stage B proves that integrating HyperVQ directly yields massive bitrate savings without altering the pre-trained latent representations. This makes the "Quick Application" highly valuable for rigid, legacy VQ systems. However, Stage C further shifts the RD frontier downwards and to the left. By allowing the network to actively optimize the feature embeddings under the guidance of the learned entropy model, joint training unlocks the true potential of Rate-Distortion-Perception control.

\section{Theoretical Derivation and Loss Formulation}
\label{sec:supp_theory}

In this section, we provide a rigorous, step-by-step mathematical derivation of the HyperVQ objective function. We aim to demonstrate the complete logical chain from the fundamental Rate-Distortion definition to the final differentiable loss term.

\subsection{Derivation of the VQ Rate Objective}
Let $x$ denote the input image and $\hat{z}_b$ denote the quantized hyper-latent at granularity $b$. Our goal is to minimize the bitrate $R_{\hat{y}_b}$ required to transmit the VQ indices $\hat{y}_b$. Let $P^\star_{b,ij}(k)$ be the true (unknown) posterior probability that the index at location $(i,j)$ is $k$, and let $P_{\theta,b,ij}(k)$ be the predicted probability parameterized by $\theta$.

The rate $R_{\hat{y}_b}$ is defined as the expected code length. We expand the cross-entropy term step-by-step:
\begin{align}
	R_{\hat{y}_b} 
	&= \mathbb{E}_{x} \left[ \sum_{i,j} H(P^\star_{b,ij}, P_{\theta,b,ij}) \right] \\
	&= \mathbb{E}_{x} \left[ \sum_{i,j} \sum_{k=1}^K - P^\star_{b,ij}(k) \log P_{\theta,b,ij}(k) \right] \\
	&= \mathbb{E}_{x} \Bigg[ \sum_{i,j} \sum_{k=1}^K - P^\star_{b,ij}(k) \log \left( \frac{P^\star_{b,ij}(k)}{P^\star_{b,ij}(k)} P_{\theta,b,ij}(k) \right) \Bigg] \\
	&= \mathbb{E}_{x} \Bigg[ \sum_{i,j} \Bigg( \underbrace{\sum_{k=1}^K - P^\star_{b,ij}(k) \log P^\star_{b,ij}(k)}_{H(P^\star_{b,ij})} \nonumber \\
	&\quad + \underbrace{\sum_{k=1}^K P^\star_{b,ij}(k) \log \frac{P^\star_{b,ij}(k)}{P_{\theta,b,ij}(k)}}_{D_{\mathrm{KL}}(P^\star_{b,ij} \| P_{\theta,b,ij})} \Bigg) \Bigg]
\end{align}

Since the entropy of the true distribution $H(P^\star_{b,ij})$ is independent of the network parameters $\theta$, minimizing the rate is strictly equivalent to minimizing the KL divergence. In our deterministic VQ framework, the "true" index $k^\star_{b,ij}$ is obtained via nearest-neighbor search:
\begin{equation}
	k^\star_{b,ij} = \arg\min_k \| (y_b)_{ij} - \mathbf{e}_k \|^2
\end{equation}
Thus, the true distribution is a Dirac delta function: $P^\star_{b,ij}(k) = \delta(k, k^\star_{b,ij})$. Substituting this into the rate equation simplifies the objective to the Negative Log-Likelihood (NLL):
\begin{align}
	R_{\hat{y}_b} 
	&\equiv \mathbb{E}_{x} \left[ \sum_{i,j} \sum_{k=1}^K - \delta(k, k^\star_{b,ij}) \log P_{\theta,b,ij}(k) \right] \\
	&= \mathbb{E}_{x} \left[ \sum_{i,j} - \log P_{\theta,b,ij}(k^\star_{b,ij}) \right] \label{eq:supp_nll}
\end{align}

\subsection{From Continuous Hyperprior to Discrete Probability}
The hyperprior network outputs the parameters of a continuous Gaussian distribution $\mathcal{N}(\mathbf{h}; \boldsymbol{\mu}_{b,ij}, \boldsymbol{\Sigma}_{b,ij})$ in the embedding space. By modeling the discrete probability mass as being proportional to the likelihood density at the quantization centroids, we convert the continuous density into a categorical distribution:
\begin{align}
	P_{\theta,b,ij}(k) 
	&\approx \frac{p(\mathbf{e}_k | \hat{z}_b)}{\sum_{l=1}^K p(\mathbf{e}_l | \hat{z}_b)} \\
	&= \frac{\exp\left( -\frac{1}{2} d^2_{\mathcal{M}}(\mathbf{e}_k, \boldsymbol{\mu}_{b,ij}) \right)}{\sum_{l=1}^K \exp\left( -\frac{1}{2} d^2_{\mathcal{M}}(\mathbf{e}_l, \boldsymbol{\mu}_{b,ij}) \right)}
\end{align}
where $d^2_{\mathcal{M}}$ is the squared Mahalanobis distance.

\subsection{Expansion under Isotropic Assumption}
To ensure computational efficiency and numerical stability, we assume an isotropic covariance $\boldsymbol{\Sigma}_{b,ij} = (\sigma_{b,ij})^2 \mathbf{I}$. This reduces the Mahalanobis distance to a scaled squared Euclidean distance. Substituting this probability model back into Eq.~(\ref{eq:supp_nll}), the loss for a single spatial location expands to:
\begin{align}
	\mathcal{L}_{rate} 
	&= - \log P_\theta(k^\star) \\
	&= - \log \left( \frac{\exp\left( - \frac{\| \mathbf{e}_{k^\star} - \boldsymbol{\mu} \|^2}{2\sigma^2} \right)}{\sum_{l=1}^K \exp\left( - \frac{\| \mathbf{e}_l - \boldsymbol{\mu} \|^2}{2\sigma^2} \right)} \right) \\
	\begin{split}
		&= - \Bigg[ \log \left( \exp\left( - \frac{\| \mathbf{e}_{k^\star} - \boldsymbol{\mu} \|^2}{2\sigma^2} \right) \right) \\
		&\quad - \log \left( \sum_{l=1}^K \exp\left( - \frac{\| \mathbf{e}_l - \boldsymbol{\mu} \|^2}{2\sigma^2} \right) \right) \Bigg] 
	\end{split} \\
	&= \underbrace{\frac{\| \mathbf{e}_{k^\star} - \boldsymbol{\mu} \|^2}{2\sigma^2}}_{\text{Anchor Alignment}} + \underbrace{\log \left( \sum_{l=1}^K \exp\left( - \frac{\| \mathbf{e}_l - \boldsymbol{\mu} \|^2}{2\sigma^2} \right) \right)}_{\text{Log-Partition (Uncertainty penalty)}} \label{eq:supp_final_expanded_loss}
\end{align}

\subsection{Centroid Approximation Bound}
\label{sec:supp_centroid_bound}
In Equation (3) of the main paper, we evaluate the density precisely at the codebook centroids $p_k \propto \mathcal{N}(\mathbf{e}_k; \boldsymbol{\mu}, \sigma^2 \mathbf{I})$ rather than performing an exact probability mass integral over the entire Voronoi cell $q_k = \int_{\mathcal{V}_k} \mathcal{N}(x; \boldsymbol{\mu}, \sigma^2 \mathbf{I}) dx$. This is a deliberate centroid-density approximation designed to maintain computational tractability in high-dimensional embedding spaces. Empirically, it provides stable, differentiable gradients without the prohibitive computational cost of high-dimensional cell integration.

To mathematically justify this approximation, we provide a theoretical bound demonstrating that it is highly accurate when codebook cells are relatively small and regular. Let the Voronoi cell $\mathcal{V}_k$ have a bounded radius $r_k$, such that for any $x \in \mathcal{V}_k$, $\|x - \mathbf{e}_k\| \le r_k$. For the isotropic Gaussian density $\phi(x) = \mathcal{N}(x; \boldsymbol{\mu}, \sigma^2 \mathbf{I})$, the variation in log-density for any $x \in \mathcal{V}_k$ compared to the centroid $\mathbf{e}_k$ can be evaluated by expanding the squared distance $\|x - \boldsymbol{\mu}\|^2 = \|(x - \mathbf{e}_k) + (\mathbf{e}_k - \boldsymbol{\mu})\|^2$. The difference is bounded by:
\begin{align}
	|\log \phi(x) - \log \phi(\mathbf{e}_k)| &= \left| -\frac{\|x - \boldsymbol{\mu}\|^2}{2\sigma^2} - \left(-\frac{\|\mathbf{e}_k - \boldsymbol{\mu}\|^2}{2\sigma^2}\right) \right| \\
	&\le \frac{\|\mathbf{e}_k - \boldsymbol{\mu}\| r_k + r_k^2 / 2}{\sigma^2}
\end{align}
This inequality formally demonstrates that when the Voronoi cells are relatively small (small $r_k$), the codebook is regular, and the predicted uncertainty spread $\sigma$ is not too small, the centroid density closely approximates the true cell mass. Mismatches predominantly occur only when the codebook cell volume is highly non-uniform or when $\sigma \to 0$. In practice, the network learns to adapt its predicted $\sigma$ and $\boldsymbol{\mu}$ to account for these quantization cells, making the empirical coding loss negligible while completely avoiding the $\mathcal{O}(D^3)$ or numerical integration costs associated with exact Voronoi mass computation.

\subsection{Gradient Analysis for Differentiability}
To demonstrate exactly how this formulation enables end-to-end differentiability, we apply the chain rule to Eq.~(\ref{eq:supp_final_expanded_loss}). The gradient with respect to the predicted mean $\boldsymbol{\mu}$ splits into two terms:
\begin{align}
	\frac{\partial \mathcal{L}_{rate}}{\partial \boldsymbol{\mu}} 
	&= \frac{\partial}{\partial \boldsymbol{\mu}} \left( \frac{\| \mathbf{e}_{k^\star} - \boldsymbol{\mu} \|^2}{2\sigma^2} \right) \nonumber \\
	&\quad + \frac{\partial}{\partial \boldsymbol{\mu}} \log \left( \sum_{l=1}^K \exp\left( - \frac{\| \mathbf{e}_l - \boldsymbol{\mu} \|^2}{2\sigma^2} \right) \right) \\
	\begin{split}
		&= - \frac{(\mathbf{e}_{k^\star} - \boldsymbol{\mu})}{\sigma^2} \\
		&\quad + \frac{\sum_{l=1}^K \exp\left( - \frac{\| \mathbf{e}_l - \boldsymbol{\mu} \|^2}{2\sigma^2} \right) \frac{(\mathbf{e}_l - \boldsymbol{\mu})}{\sigma^2}}{\sum_{j=1}^K \exp\left( - \frac{\| \mathbf{e}_j - \boldsymbol{\mu} \|^2}{2\sigma^2} \right)}
	\end{split} \\
	\begin{split}
		&= - \frac{(\mathbf{e}_{k^\star} - \boldsymbol{\mu})}{\sigma^2} \\
		&\quad + \sum_{l=1}^K \underbrace{\left( \frac{\exp( - \frac{\| \mathbf{e}_l - \boldsymbol{\mu} \|^2}{2\sigma^2} )}{\sum_{j=1}^K \exp( - \frac{\| \mathbf{e}_j - \boldsymbol{\mu} \|^2}{2\sigma^2} )} \right)}_{P_\theta(l)} \frac{(\mathbf{e}_l - \boldsymbol{\mu})}{\sigma^2}
	\end{split} \\
	&= \frac{1}{\sigma^2} \left( \sum_{l=1}^K P_\theta(l) (\mathbf{e}_l - \boldsymbol{\mu}) - (\mathbf{e}_{k^\star} - \boldsymbol{\mu}) \right)
\end{align}
The final gradient acts as an elegant, differentiable force field: it pulls the predicted mean $\boldsymbol{\mu}$ towards the target codebook anchor $\mathbf{e}_{k^\star}$ (first term), while pushing it away from all other codebook vectors $\mathbf{e}_l$, weighted by their currently predicted probabilities $P_\theta(l)$ (second term).

\section{Detailed Architecture of HVQ-CGIC}
\label{sec:supp_arch}

\begin{figure*}[ht]
	\centering
	\includegraphics[width=1.0\linewidth]{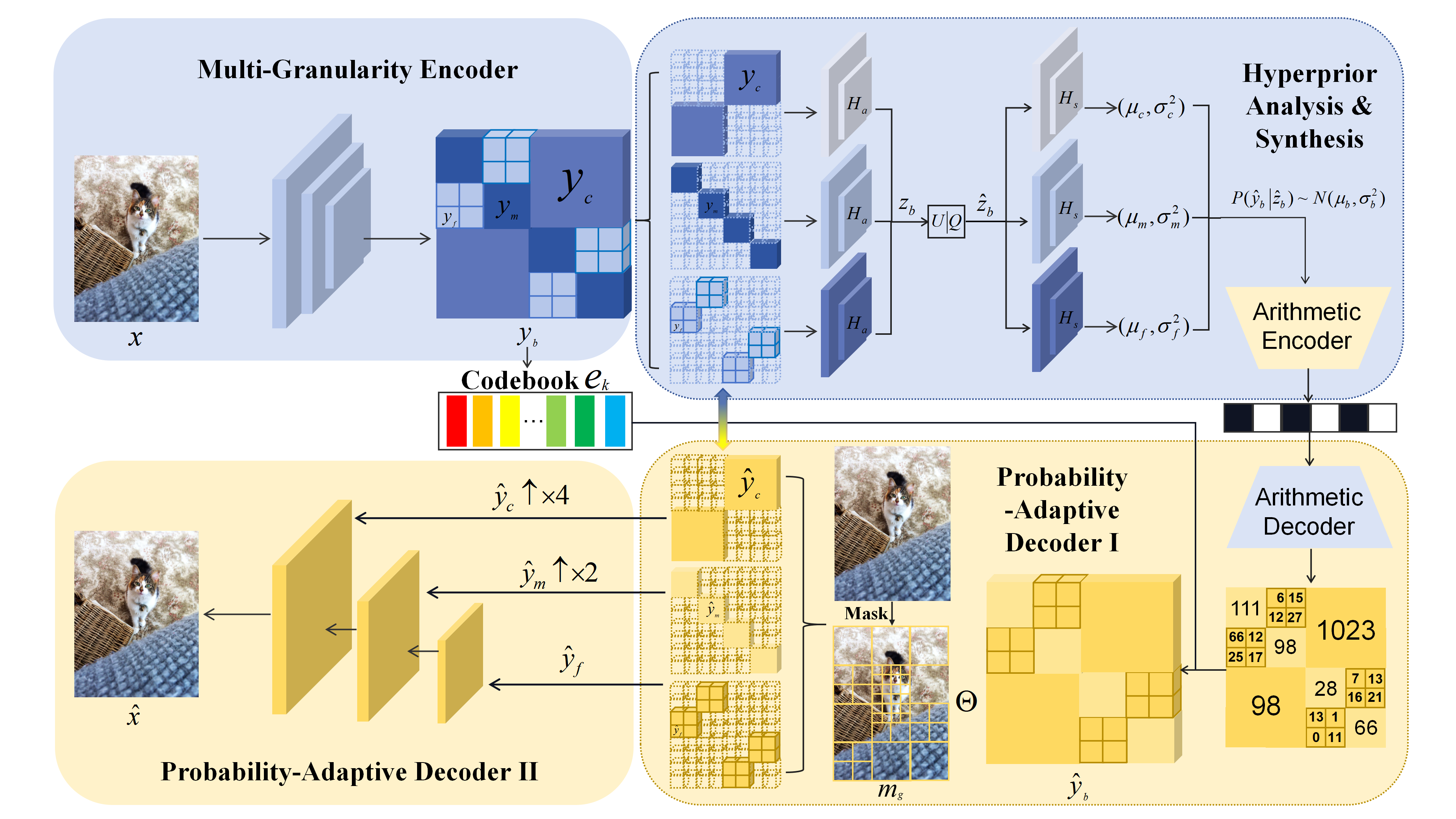}
	\caption{\textbf{Detailed Flowchart of HVQ-CGIC.} The framework employs a Multi-Granularity Encoder and shares a unified codebook across granularities. The HyperVQ module robustly predicts distributions for the sparse, dynamically masked indices. $\otimes$ represents element-wise multiplication with spatial routing masks.}
	\label{fig:supp_arch}
\end{figure*}

While HyperVQ serves as a universal plug-in for any VQ-based codec, we specifically instantiate it on a controllable multi-granularity scenario to rigorously prove its robustness. The complete flowchart of this instantiation, termed HVQ-CGIC, is illustrated in Figure~\ref{fig:supp_arch}.

\textbf{The Challenge of Arbitrary Rate Control.} Standard VQ codecs typically process fixed-scale, full-resolution latent maps. In contrast, the multi-granularity framework dynamically routes features across multiple scales (coarse, medium, and fine) to achieve continuous rate adaptation. This mechanism inherently produces sparse, dynamically masked, and partial feature maps. Traditional statistical entropy models fail to adapt to these non-uniform distributions efficiently.

\textbf{Overall Architecture Design.} As depicted in Figure~\ref{fig:supp_arch}, the HVQ-CGIC framework begins with a \textbf{Multi-Granularity Encoder} that extracts hierarchical representations. These features are then quantized using a unified, shared \textbf{Codebook $\mathcal{E}$}. 

To tackle the sparse latent challenge, our \textbf{Hyperprior Analysis \& Synthesis} module processes the quantized hyper-latents $\hat{z}_b$ to dynamically predict the continuous Gaussian parameters $(\boldsymbol{\mu}, \sigma^2)$ for the active regions defined by the routing masks. Specifically, the \textbf{Hyper Synthesis block} is structurally positioned to directly parameterize the probability mapping, seamlessly adapting to the varying scales and accurately modeling the probabilities of the masked indices.

Finally, the \textbf{Probability-Adaptive Decoders} utilize these multi-scale indices, modulated by the spatial masks $m_i$, to reconstruct the final image $\hat{x}$. By unifying the multi-scale VQ embeddings under a single hyperprior probability model, HVQ-CGIC unlocks precise Rate-Distortion optimization even for arbitrarily masked, sparse latent representations.

\section{Comparison with Traditional High-Order Context Modeling}
\label{sec:supp_context}

\begin{figure}[htbp]
	\centering
	\includegraphics[width=1.0\linewidth]{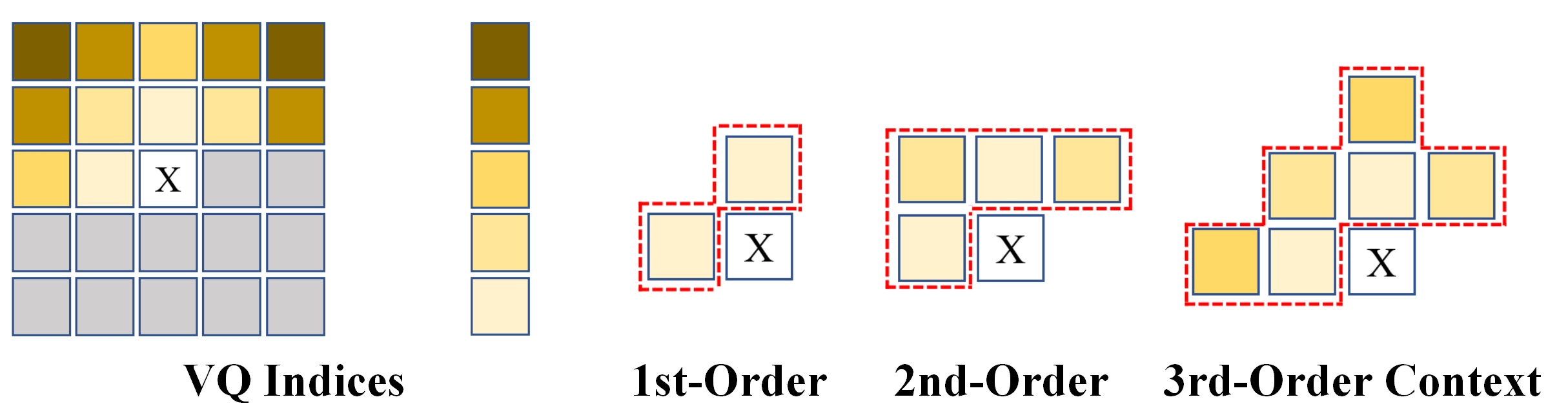}
	\caption{Illustration of spatial context designs. Building upon standard statistical entropy coding, these high-order context strategies perform adaptive arithmetic coding by dynamically updating symbol encoding frequencies in real-time based on neighboring test data.}
	\label{fig:supp_context_design}
\end{figure}

A traditional method for improving baseline statistical entropy coding relies on designing adaptive arithmetic coders enriched with high-order spatial contexts (as illustrated in Figure~\ref{fig:supp_context_design}). These strategies incrementally build upon basic probability tables by dynamically updating symbol encoding frequencies in real-time based on local neighbor combinations observed during inference.

However, as presented in Table~\ref{tab:context_ablation}, manually tracking index-based spatial contexts demonstrates severe diminishing marginal returns. While a 3rd-order context does improve the BPP marginally over the baseline Control-GIC (from 0.594 down to 0.583), the sequential nature of maintaining and querying these complex frequency states balloons the encoding time to an unacceptable $145.2$ ms. 

To evaluate our method against these traditional approaches, we present two variants of our model corresponding to the progressive integration protocols detailed in Appendix~\ref{sec:supp_training}. Specifically, HVQ-CGIC (BPP) refers to the model after Stage B (Rapid Hyperprior Adaptation), where the hyperprior is trained exclusively to minimize the bitrate while the VQ backbone remains frozen. Conversely, HVQ-CGIC (RD) denotes the model after Stage C (Joint RD Fine-tuning), representing the fully co-adapted, end-to-end optimized framework that balances both rate and distortion.

\begin{table*}[htbp]
	\centering
	\renewcommand{\arraystretch}{1.3}
	\setlength{\tabcolsep}{3pt}
	\caption{Ablation study on the Kodak dataset (downsampling factor 4). Percentages in \textcolor{teal}{green} indicate relative improvements over the theoretical uncompressed limit (No Compression). The Enc. and Dec. times represent the batched inference latency (sum of hyperprior network inference and arithmetic coding time), highlighting the massive throughput advantage of parallel execution.}
	\begin{tabular}{l l l c c}
		\toprule
		\textbf{Method}  & \textbf{BPP} ($\downarrow$) & \textbf{LPIPS} ($\downarrow$) &\textbf{Enc.} (ms) &\textbf{Dec.} (ms) \\
		\midrule
		No Compression  & 0.625 & -  & - & - \\
		Control-GIC\cite{li2024once} & 0.594 (\textcolor{teal}{-4.9\%})  & 0.0248 & 8.43 & 18.1 \\
		\midrule
		1st-Order Context & 0.587 (\textcolor{teal}{-6.0\%}) & 0.0248 & 4.66 & 9.04 \\
		2nd-Order Context & 0.584 (\textcolor{teal}{-6.5\%}) & 0.0248 & 32.9 & 48.4 \\
		3rd-Order Context & 0.583 (\textcolor{teal}{-6.7\%}) & 0.0248 & 145.2 & 190.2 \\
		\midrule
		HVQ-CGIC (RD) & 0.530 (\textcolor{teal}{-15.2\%}) & \textbf{0.0197}  & 2.78 & 4.98 \\
		HVQ-CGIC (BPP) & \textbf{0.493} (\textcolor{teal}{\textbf{-21.1\%}}) & 0.0278 & \textbf{2.71} & \textbf{4.17} \\
		\bottomrule
	\end{tabular}
	\label{tab:context_ablation}
\end{table*}

In sharp contrast to traditional methods, our HyperVQ approach implicitly learns an ultra-high-dimensional context representation via its lightweight parametric network, entirely bypassing the need for sequential frequency table updates. Consequently, our Stage B model, HVQ-CGIC (BPP), extracts far more latent correlation than traditional 3rd-order contexts, achieving an impressive $0.493$ BPP. Furthermore, the parallel capabilities of our Probability Estimation Engine (discussed in Appendix~\ref{sec:supp_algo}) drastically reduce the batched encoding latency to just $2.71$ ms. Meanwhile, the fully fine-tuned HVQ-CGIC (RD) achieves an optimal balance, significantly improving the perceptual quality (LPIPS drops to 0.0197) while maintaining a highly competitive bitrate and ultra-low inference latency.

\section{Analysis: Why Not a Direct K-Way Classifier?}
\label{sec:supp_kway}

When attempting to establish content-adaptive entropy modeling for discrete Vector Quantization (VQ) indices, the most intuitive approach might be to utilize a neural network as a direct K-way classifier (where $K$ is the codebook size) to output a categorical distribution directly. However, in practice, we found that such non-parametric probability models fail to converge efficiently or scale properly. 

\textbf{1. Parameter Explosion ($\mathcal{O}(K)$ vs $\mathcal{O}(D)$):}
A direct K-way classifier requires outputting a logit for every single codebook entry at every spatial location. For a latent map of size $W \times H$, the output tensor shape is $W \times H \times K$. For a standard codebook size of $K=1024$, this represents a massive prediction space. In contrast, our HyperVQ module is a \textbf{parametric model} that only requires predicting the parameters $(\boldsymbol{\mu}, \sigma)$ of a Gaussian distribution in the $D$-dimensional continuous embedding space. Assuming $D=4$, our network only outputs a tensor of shape $W \times H \times (D+1)$. This means the K-way classifier requires over $\mathbf{200\times}$ more parameters at the final prediction layer, making it highly memory-intensive and prone to overfitting.

\textbf{2. Codebook-Size-Independent Scalability:}
Modern advancements in VQ-based generation are rapidly moving towards ultra-large codebooks ($K = 100,000$ or larger). A K-way classifier becomes mathematically intractable to optimize for such large vocabularies due to the vanishing gradients in the immense Softmax calculation. Our HyperVQ framework fundamentally bypasses this limit: since the network only predicts the $D$-dimensional mean feature $\boldsymbol{\mu}$, its complexity is entirely \textbf{codebook-size-independent}.

\section{Entropy Coding and Spatial Bit Usage Analysis}
\label{sec:supp_entropy_bit}

\begin{figure}[htbp]
	\centering
	\includegraphics[width=0.8\linewidth]{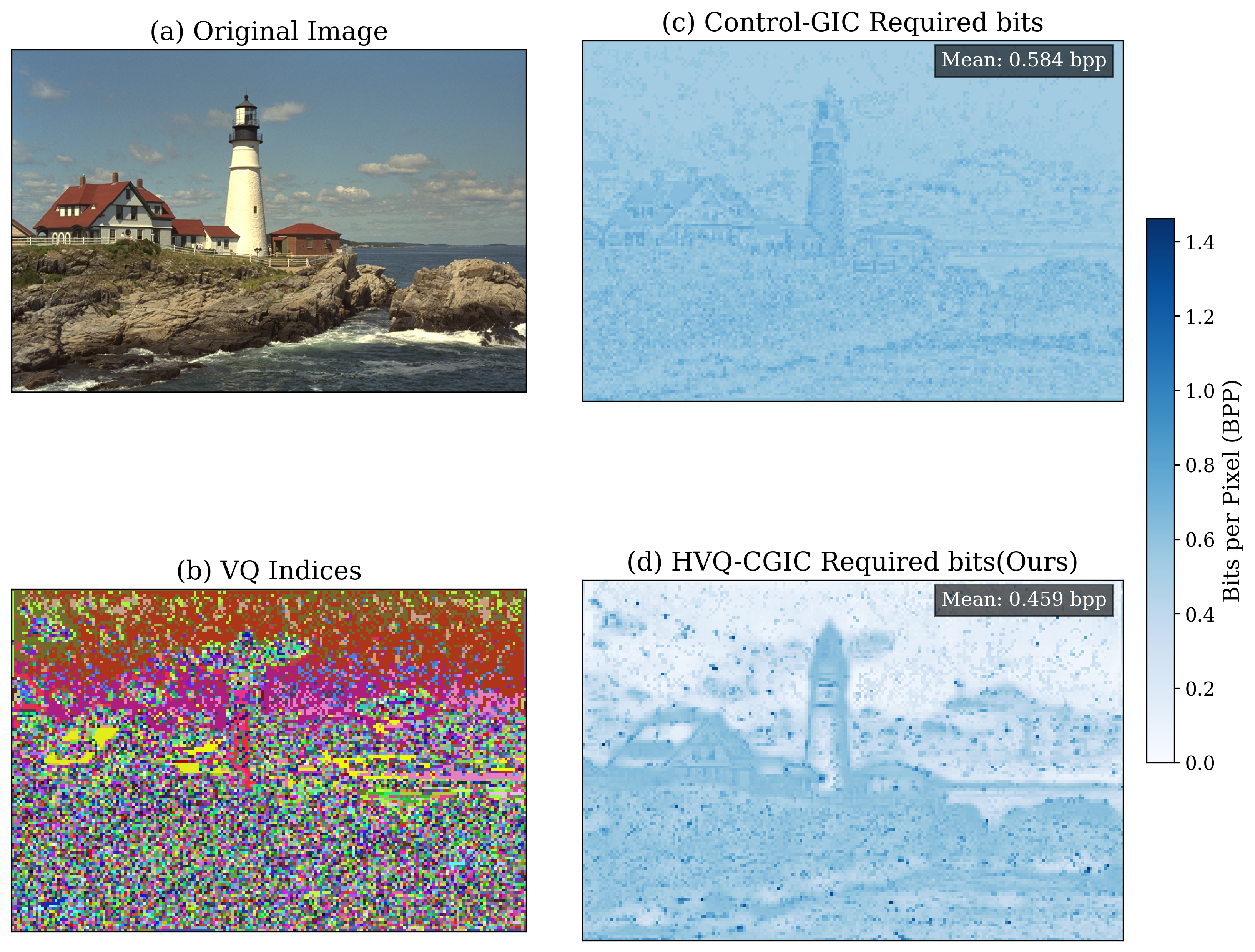}
	\caption{\textbf{Visualization of spatial bit usage distribution on $Kodim21$.} HVQ-CGIC almost exhausts every bit of coding redundancy in the latent feature space.}
	\label{fig:supp_bit_distribution}
\end{figure}

To intuitively understand why HyperVQ excels under dynamic constraints, we visualized the spatial bit usage distribution of $Kodim21$ at a $\times 4$ downsampling ratio compared to the Control-GIC baseline. As shown in Figure~\ref{fig:supp_bit_distribution}, our framework almost exhausts every bit of coding redundancy in the latent feature space. By capturing the underlying structural correlations via the embedding-space Gaussian prior, HVQ-CGIC saves over 21.4\% bitrate compared to Control-GIC on this single image, without any loss in textural fidelity.

\section{Extended Qualitative Visual Comparisons}
\label{sec:supp_visuals}

This section provides extensive visual comparisons to demonstrate the perceptual superiority of HVQ-CGIC in complex scenes, high-frequency textures, and structured regions. We compare our framework against SOTA baselines (HiFiC, Control-GIC, CDC) on the Kodak, DIV2K, and CLIC 2020 datasets.

\begin{figure*}[htbp]
	\centering
	\includegraphics[width=0.95\linewidth]{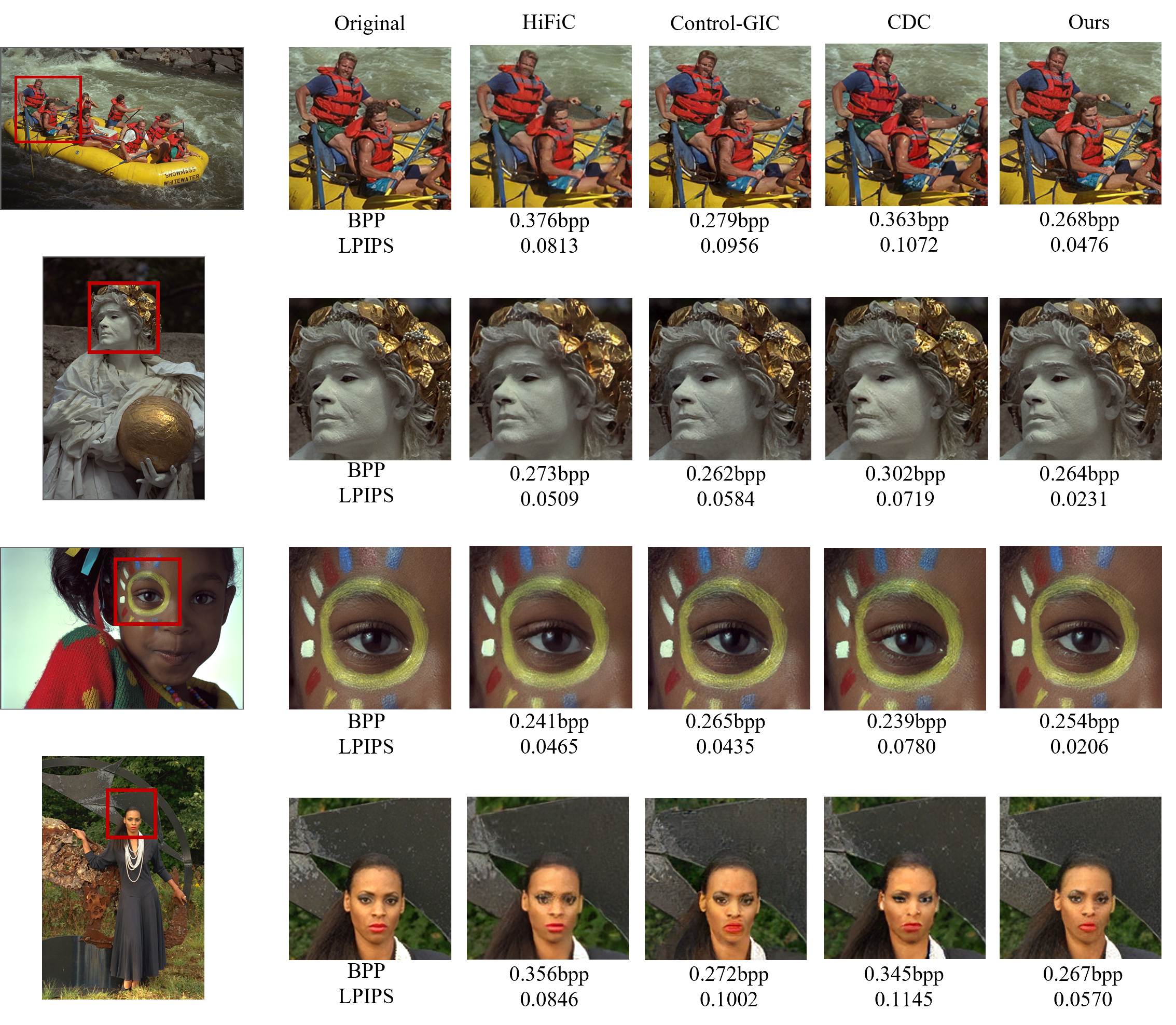}
	\caption{\textbf{Qualitative Comparison on Kodak Dataset.} Best viewed when zoomed in. HVQ-CGIC preserves sharp structural edges and suppresses artifacts better than baselines in structured regions.}
	\label{fig:supp_kodak_recon}
\end{figure*}

\textbf{Structure Preservation (Kodak).} Figure \ref{fig:supp_kodak_recon} visualizes performance on structured regions (e.g., text, geometric lines). HVQ-CGIC preserves sharp edges and accurate geometry, avoiding the blurring or ringing artifacts observed in baselines. 

\begin{figure*}[htbp]
	\centering
	\includegraphics[width=0.95\linewidth]{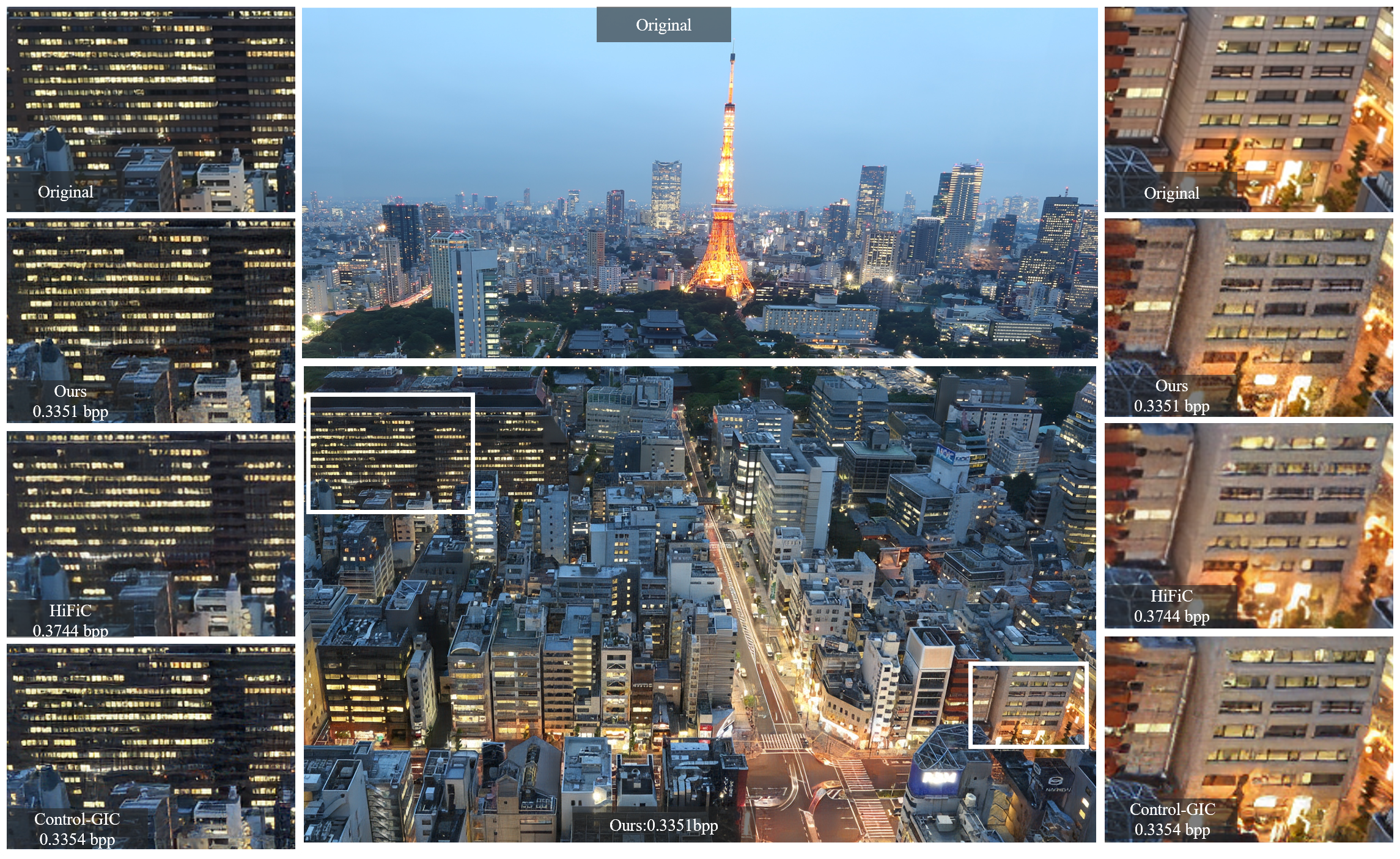}
	\caption{\textbf{Qualitative Comparison on DIV2K Dataset.} Comparison of fine-grained texture reconstruction.}
	\label{fig:supp_div2k_recon}
\end{figure*}

\textbf{Texture Synthesis (DIV2K).} Figure \ref{fig:supp_div2k_recon} demonstrates the reconstruction of dense, stochastic textures (e.g., foliage, urban details). While previous VQ methods tend to over-smooth fine grain, HVQ-CGIC faithfully synthesizes high-frequency details, yielding a photorealistic appearance.

\begin{figure*}[htbp]
	\centering
	\includegraphics[width=0.95\linewidth]{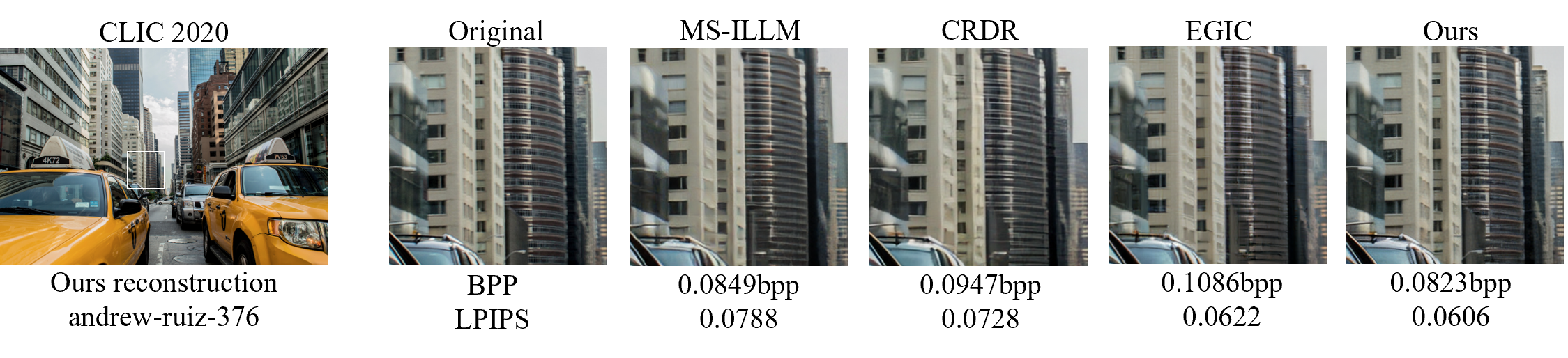}
	\caption{\textbf{Qualitative comparison at extremely low bit rates on CLIC 2020.} HVQ-CGIC reconstructs faithful textures with far fewer artifacts.}
	\label{fig:supp_clic_recon_extra}
\end{figure*}

\begin{figure*}[htbp]
	\centering
	\includegraphics[width=0.95\linewidth]{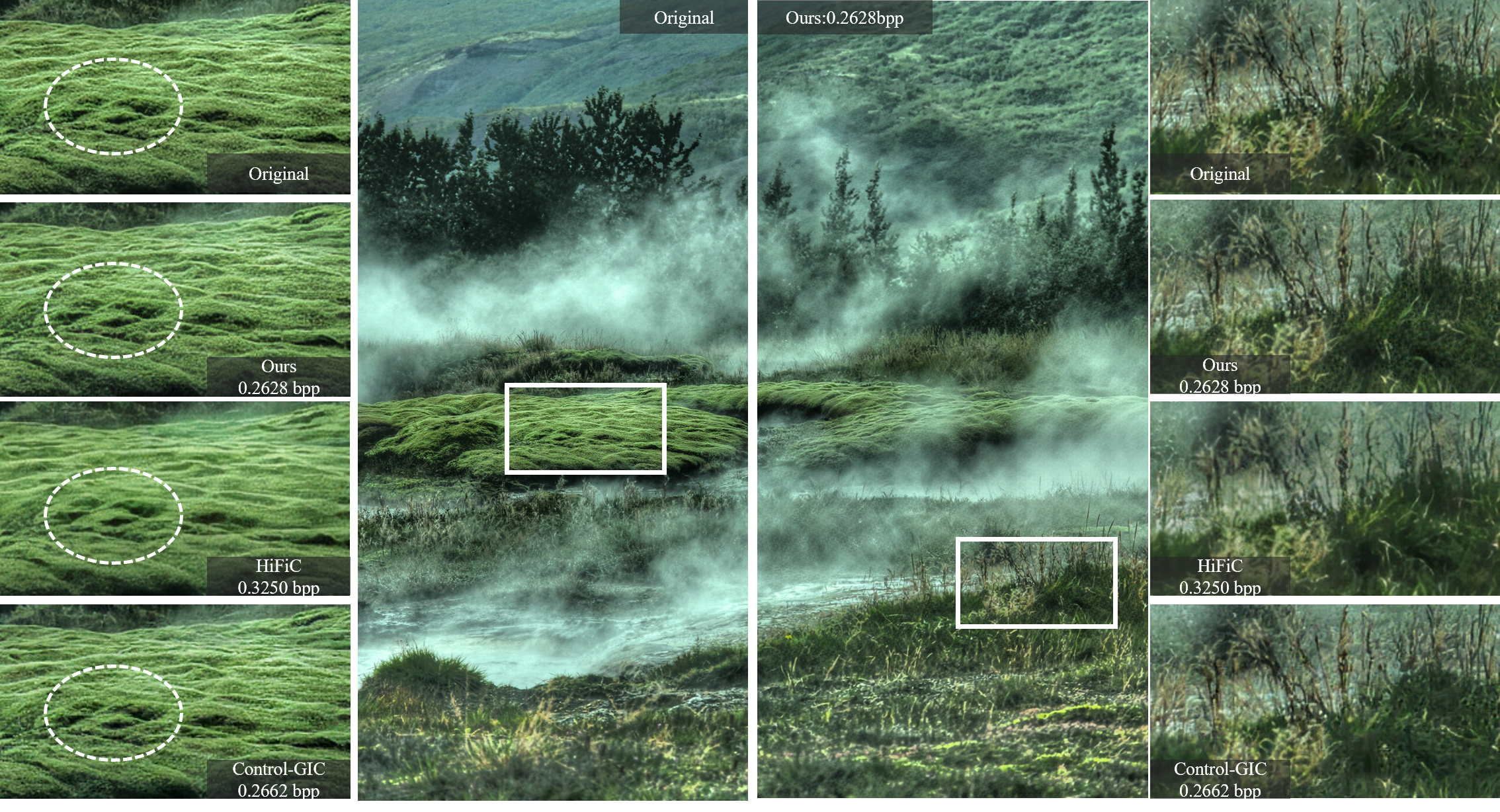}
	\caption{\textbf{Qualitative Comparison on CLIC 2020 Dataset.} Comparison of perceptual quality in complex scenes.}
	\label{fig:supp_clic_recon}
\end{figure*}

\textbf{Complex Scenes (CLIC 2020).} In complex real-world scenes (Figures \ref{fig:supp_clic_recon_extra} and \ref{fig:supp_clic_recon}), HVQ-CGIC exhibits superior global consistency. It strongly suppresses generative artifacts and color shifts, maintaining fidelity to the original content supported by the precise probability estimation of the VQ indices.

\clearpage


\end{document}